\documentclass[journal]{IEEEtran}
\usepackage{graphicx}
\usepackage{amsmath, amsthm, amssymb}
\usepackage{color}
\usepackage{multirow}
\usepackage{times}
\usepackage{subfig}
\usepackage{cite}
\usepackage{makecell}
\usepackage{stfloats}

\usepackage{algorithm} % Z. Song 4/8/2020
\usepackage{algorithmicx} % Z. Song 4/8/2020
\usepackage[noend]{algpseudocode} % Z. Song 4/8/2020
\usepackage{comment}
\allowdisplaybreaks

\newcommand{\tabincell}[2]{\begin{tabular}{@{}#1@{}}#2\end{tabular}}

\newtheorem{remark}{Remark}

\newtheorem{theorem}{Theorem}
\newtheorem{property}{Property}
\newtheorem{assumption}{Assumption}

\newtheorem{proposition}{Proposition}

\newtheorem{definition}{Definition}

\ifCLASSINFOpdf
\else
\fi
\begin{document}
\title{Resilient Time-Varying Formation Tracking for Mobile Robot Networks under Deception Attacks on Positioning}

\author{Yen-Chen~Liu, Kai-Yuan Liu, and Zhuoyuan Song
\thanks{This work was supported by the Ministry of Science and Technology (MOST), Taiwan, under grants MOST 110-2636-E-006-005 and  MOST 109-2636-E-006-019. Z.~Song was partially supported by the U.S.  National Science Foundation under awards CISE/IIS–2024928 and OIA-2032522.}
\thanks{Y.-C. Liu and K.-Y. Liu are with the Department of Mechanical Engineering, National Cheng Kung University, Tainan, Taiwan (e-mail: \texttt{yliu@mail.ncku.edu.tw,k242424123@gmail.com}).}% <-this % stops a space
\thanks{Z.~Song is with the Department of Mechanical Engineering, University of Hawai`i at M\={a}noa, Honolulu, HI, USA (e-mail: \texttt{zsong@hawaii.edu}).}
%\thanks{K.-Y. Liu is with the Department of Mechanical Engineering, National Cheng Kung University, Tainan 70101, Taiwan (e-mail: \texttt{k242424123@gmail.com}).}
}

\maketitle

\begin{abstract}
%To be finished.
%\red{We need to come up with a bullet list of novelties and original contributions.}
%\begin{itemize}
%    \item A resilient position estimation system automatically switches in between external global positioning and decentralized cooperative localization [ZS]
%    \item An external positioning signal fault detection and isolation scheme based on Kullback-Leibler divergence [ZS]
%    \item A communication-free cooperative localization approach based on extended information filters [ZS]
%    \item A time-varying formation tracking control for mobile robot network to provide resilience utilizing redundancy [YC]
%    \item An illustration of distinction between resilience and robustness of networked robotic system [YC]
%    \item Analysis and design of positioning attack with performance index and resilience operation for mobile robot network in time-varying formation tracking control [YC].
%    \item Designs of performance index and resilience measure [YC].
%    \item Design of resilience index and the resilience leverages techniques synthesizing the formation control and cooperative %localization.
%\end{itemize}
This paper investigates the resilient control, analysis, recovery, and operation of mobile robot networks in time-varying formation tracking under deception attacks on global positioning. 
Local and global tracking control algorithms are presented to ensure redundancy of the mobile robot network and to retain the desired functionality for better resilience.
Lyapunov stability analysis is utilized to show the boundedness of the formation tracking error and the stability of the network under various attack modes.
A performance index is designed to compare the efficiency of the proposed formation tracking algorithms in situations with or without positioning attacks.
Subsequently, a communication-free decentralized cooperative localization approach based on extended information filters is presented for positioning estimate recovery where the identification of the positioning attacks is based on Kullback–Leibler divergence.
A gain-tuning resilient operation is proposed to strategically synthesize the formation control and cooperative localization for accurate and rapid system recovery from positioning attacks. 
The proposed methods are tested using both numerical simulation and experimental validation with a team of quadrotors. 
%The redundancy, robustness, resourcefulness, and rapidity of the proposed resilient control are discussed for time-varying formation tracking with mobile robot networks.
\end{abstract}


% \fi

%\IEEEpeerreviewmaketitle

%Use {\bf hardly} instead of {\bf can not}.

%{\bf The advantage of~\cite{schwager2009decentralized} is decentralized, because the method proposed in this paper is centralized.}

%%%%%%%%%%%%%%%%%%%%%%%%%%%%%%%%%%%%%%%%%%%%%%%%%%%%%%%%%%%%
\section{Introduction}\label{sec:intro}

Formation tracking and control are significant research topics for networked mobile robots such as unmanned ground vehicles~\cite{Desai01TRA,Sun09TRO} and autonomous aerial~\cite{StipanovicD:04a,DongX:15a} or underwater vehicles~\cite{CuiR:10a,DasB:16a}. 
Owing to the benefits of efficiency, redundancy, robustness, scalability, flexibility, and reliability, maintaining a formation during the entire mission in mobile robot networks offers many advantages across various applications such as surveillance, drag reduction, source seeking, environmental sampling, search and rescue operations, aerial refueling, cooperative transportation, and closed-formation flight~\cite{Guillet14RAM,Giulietti00CSM,DeVries16JGCD,Dai17TIE, Loria16TCST, Liang16TAC, Yu19IJC, Wang19IJRNC,fu2020distributed,li2021adaptive}.
Although several formation-control approaches have been extensively studied including leader–follower, virtual structure, behavior-based, and consensus-based techniques, a common assumption in such studies has been the availability of inter-agent communication with accurate agent position information.
In real-world applications, data transmission over cyber networks and information shared between agents could become susceptible to not only adverse and malicious threats but also attacks that would crucially compromise or even destroy a networked robotic system~\cite{BijaniS:14a,GilS:17a}.

\begin{figure}
   \centering
\hspace{-.1in}   \includegraphics[width=3.56in]{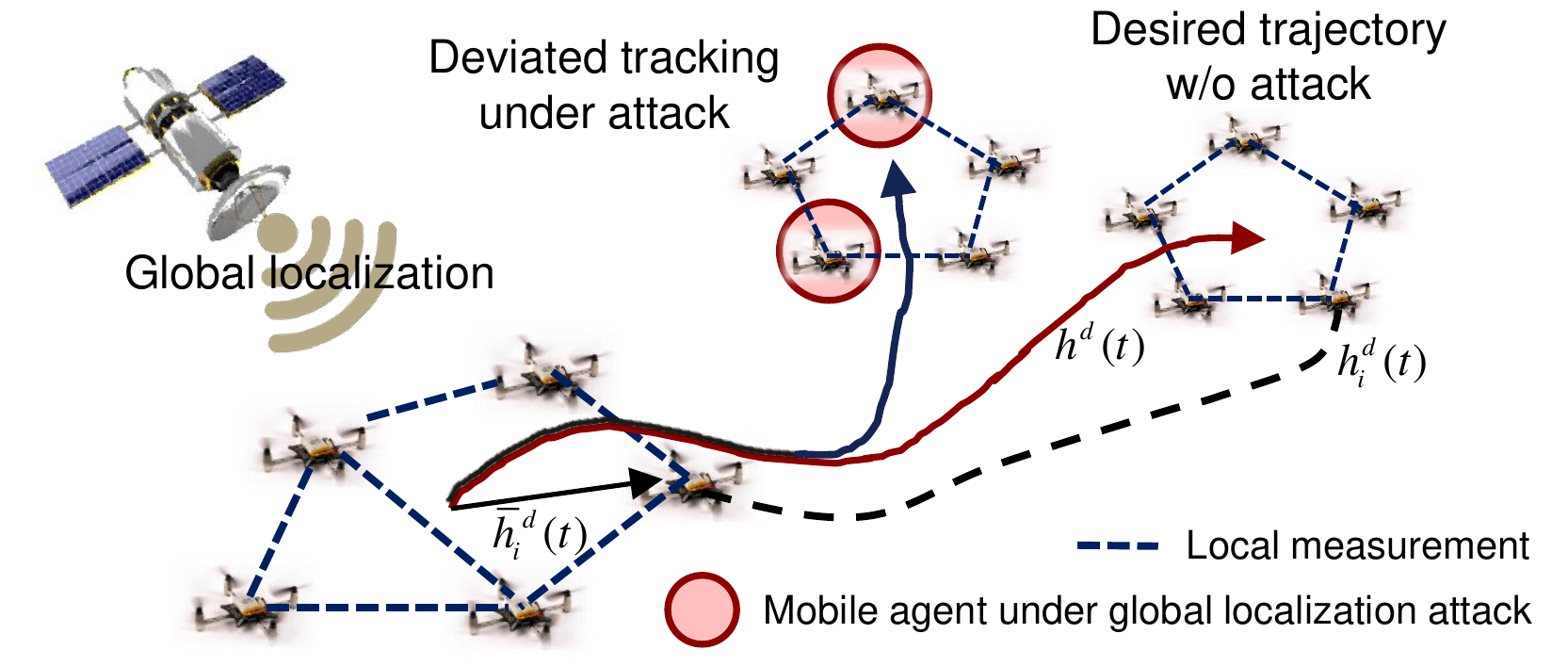}
   \caption{An illustration of time-varying formation tracking in mobile robot networks under global positioning attacks. 
   %The mobile robots form a desired time-varying formation described by $\bar{h}_i^d(t)$ with respect to a local coordinate attached to the center of the formation. The formation is designed to track a global desired time-varying trajectory $h^d(t)$ (red trajectory), where the desired trajectories for the individual robot is obtained as $h_i^d(t)=h^d(t)+\bar{h}_i^d(t)$. If the mobile robot network is subject to global positioning attacks, it would deviate from the desired formation and trajectory (blue trajectory).
   }
\label{fig:illustration}
\end{figure}

Although multi-agent systems are advantageous in various tasks with redundancy and scalability, the issues of actuator faults, fallible robots, and vulnerable communications have recently attracted significant attention~\cite{Minelli20RAS,Zhu19TRO,Modares20TCYB,Meng16AUTO,Wang19IJRNC,petrillo2021secure,yang2021attacks}. 
%The robustness of the network graphs in the presence of compromised robots was presented by~\cite{Guerrero17RAL} and subsequently applied to construct a resilient robot formation. 
%To deal with the consensus in the direction of motion, dynamic connectivity management has been proposed with a resilience threshold to guarantee the convergence of consensus for mobile robot teams under non-cooperative robots~\cite{Saulnier17RAL}. 
Connectivity preservation with respect to robot failures was addressed by~\cite{Minelli20RAS}, where a self-optimization of resilient topologies and an experimental validation were provided. In \cite{Modares20TCYB}, a distributed $H_\infty$ controller was designed to ensure resilience and robustness for leader–follower systems under faulty and malicious attacks on sensors and/or actuators. The problem of multi-agent systems in achieving a fixed formation under mismatched compasses, which would lead to distortion, has been studied with estimation and compensation algorithms~\cite{Meng16AUTO}. In addition to mismatched compasses, an approach using relative position information has been implemented in cooperative localization to improve localization accuracy in communication among agents~\cite{Zhu19TRO}.

%\cite{Defoort08TIE}, \cite{He17IJC}, \cite{Wang19TCNS}, \cite{Dong16CEP}, \cite{Yu19IJC}, \cite{Chen13CTA}, \cite{Turpin12AR}, \cite{DeVries16JGCD}, \cite{Jasim18TCST}, \cite{Beard01TCST}, \cite{Liu20TCYB}, \cite{Chen10IJRR}, \cite{Sun09TRO}, \cite{Rahimi14RAS}, \cite{Dasdemir14IJC}, \cite{Wang19IJRNC}, \cite{Loria16TCST},  \cite{Lopez20TRO},  \cite{Liu20ISA},  \cite{Liang16TAC},  \cite{Meng16AUTO},  \cite{Yang18SCL},  \cite{Zhu19TRO},  \cite{Nguyen20TRO},  \cite{Saldana19DARS},  \cite{Guerrero17RAL},   \cite{Saulnier17RAL},   \cite{Minelli20RAS}

%{\bf Par-4: Address resilient control in multiagent systems.}

%{\bf Par-5: Address fault-tolerance and robust control in time-varying formation control.}

Accurate localization is a crucial prerequisite for the control and coordination of multi-robot systems. Although decentralized multi-agent coordination methods tend to alleviate the system-wide disruption caused by localization failures of individual agents, the resulting control error eventually leads to sub-optimal performance across the entire network. In many applications, such as underwater robotics and indoor navigation, it is challenging or infeasible to provide constant global positioning information to all agents in a network~\cite{Bahr:09a,Song:13b,PaullL:14a}. When computational resources allow, and the environment is structured, persistently reliable localization can be achieved through simultaneous localization and mapping~\cite{SaeediS:16a,CadenaC:16a}. Cooperative localization is often applied to enable inter-agent observation and information fusion, thereby reducing the localization error propagation and maximizing the utilization of locally available global positioning information~\cite{MourikisAI:06a,WymeerschH:09a,KiaSS:16a}. Nonetheless, it is often assumed that inter-agent communication is available, making the system susceptible to communication failures and security complications.

In this paper, we study the time-varying formation control (TVFC) of a mobile robot network in the absence of inter-agent communication, thereby mitigating potential vulnerabilities to global positioning information attacks or failures. The mobile robot network is controlled to track prescribed formations by utilizing global positioning information and local inter-agent relative displacements obtained from proximity and/or bearing sensors. Although certain cyber-attacks can be avoided by adopting a communication-free framework, attacks on global positioning systems could cause catastrophic failures in multi-agent formation control. 
%To retain the desired functionality of such systems under positioning attacks or failures, a communication-free multi-modal localization system is presented.
To retain the desired functionality of such systems under positioning attacks or failures, a fault detection and isolation scheme based on Kullback-Leibler (KL) divergence is proposed for identifying positioning attacks.
Subsequently, a communication-free decentralized cooperative localization approach based on extended information filters (EIF) is proposed to enhance the resilience of the mobile robot network in TVFC.
The system is designed using a resilient gain-tuning formation-control approach to enhance the resilience and robustness of the multi-agent systems in the presence of adverse effects.
Systematic studies and analyses on the proposed approach for multi-robot formation tracking were performed.
To provide a deeper understanding of the resilience of a mobile robot network in TVFC, the resilience triangle from the performance index of TVFC is investigated.
Results from numerical simulations and experimental implementations on a group of quadrotor flying robots are presented to demonstrate system efficiency and performance in terms of improving the resilience of mobile robot networks.

The main contributions of this paper are summarized as follows:
\begin{enumerate}
\item TVFC of mobile robot networks is studied under a hierarchical architecture, with the design of the performance index representing the local and global tracking efficiency.
\item The adversarial effects of three positioning information attacks, i.e., additive attack, hybrid attack, and unstable attack, on the performance of formation control are investigated through Lyapunov stability analysis.
%\item Most of the aforementioned approaches rely heavily on communication and data-exchange between mobile agents. A communication-free cooperative localization approach is presented using EIF to recover the position of compromised agents and identify the agents under adversarial attack.
\item A resilient localization system is presented that identifies positioning attacks based on the KL divergence in fusing global positioning information with inertial navigation, and maintains consistent localization for the compromised agents using a communication-free, decentralized cooperative localization method based on EIF.
\item Resilient operation utilizing the results from cooperative localization is demonstrated to mitigate the adverse impacts on a mobile robot network subjected to global positioning attacks.
\item Experimental validation using a group of quadrotor flying robots is presented to demonstrate the efficacy and efficiency of the proposed resilient methods for TVFC of multi-robot systems.
\end{enumerate}

The remainder of this paper is organized as follows. Section~\ref{sec:prob} addresses the modeling, sensory graph, and problem formulation of TVFC for a mobile robot network. Section~\ref{sec:main} presents a theoretical analysis of the mobile robot network and the effects of malicious attacks on the performance of the TVFC. The cooperative localization, attack detection, position recovery, and resilient operation are discussed in Section~\ref{sec:CoL}. Section~\ref{sec:exp} presents an experimental validation of the proposed networked robot system. Section~\ref{sec:conclu} provides a discussion of the proposed resilient robotic systems and summarizes possible future directions on TVFC for mobile robot networks.

%%%%%%%%%%%%%%%%%%%%%%%%%%%%%%%%%%%%%%%%%%%%%%%%%%%%%%%%%%%%

\section{Preliminaries and Problem Formulation}\label{sec:prob}

\subsection{Robot Modeling and Sensory Graph}

Consider a mobile robot network composed of $N\geq 3$ dynamically controlled and fully-actuated mobile agents described as
\begin{eqnarray}\label{eq:dyn:rob}
&&\hspace{-.2in} \ddot x_i=u_i, \quad i=1,\ldots, N,
\end{eqnarray}
where $x_i\in\mathcal Q$ and $u_i\in \mathcal U_i$ with $\mathcal Q \subset \mathbb R^n$ as the compact set of the state space and $\mathcal U_i \subset \mathbb R^n$ as the admissible control set of $u_i$.

The time-varying formation tracking for mobile robot network is studied in this paper in the absence of inter-robot communication so that we have the next assumption.
%Therefore, the capacity of local measurement is required so that we have the next assumption.

\begin{assumption}\label{assump:local:mea}
The mobile agents in the robot network are equipped with range and bearing sensors that can obtain reliable relative position with respect to their neighbors.
This so-called local displacement measurement is mutual, i.e., the $i^{th}$ robot can obtain the measurement to the $j^{th}$ robot, and vice versa.
\end{assumption}

The graph theory~\cite{godsil} is utilized to describe the displacement measurement topologies (sensory graph) among $N$ agents in the network. % which is denoted by the interconnection graph $\mathcal G$.
For an interconnected graph $\mathcal G(\mathcal W)$, the vertex set is given as $\mathcal V(\mathcal G)$, the edge set is denoted by $\mathcal E(\mathcal G)\in \mathcal V(\mathcal G)\times \mathcal V(\mathcal G)$, and the weight matrix $\mathcal W=\{w_{ij}\}$ denotes the weighting for each of the edges in $\mathcal E(\mathcal G)$.
With Assumption~\ref{assump:local:mea} and the graph describing the sensory topology, $\mathcal G(\mathcal W)$ is undirected such that $w_{ij}=w_{ji}$.
The interconnection between the $N$ mobile robots in the network can be described by the weighted Laplacian matrix $L(\mathcal G)\in \mathbb R^{N\times N}$~\cite{Book10Egerstedt,Minelli20RAS} defined as
$
L(\mathcal G)=D(\mathcal G)-A(\mathcal G),
$
where $D(\mathcal G)$ is the degree matrix of $\mathcal G$, and $A(\mathcal G)$ is the corresponding adjacency matrix with entries $a_{ij}=w_{ij}$.
The diagonal terms of the Laplacian matrix are given as $[L(\mathcal G)]_{ii}=\sum_{j=1}^N w_{ij}$, and the off-diagonal elements of the Laplacain matrx are given as $[L(\mathcal G)]_{ij}=-w_{ij}$ for $i\neq j$.

To achieve formation tracking for the mobile robot network, the graph $\mathcal G$ should be connected so that the robots can obtain the inter-robot displacement.
The Laplacian matrix of an undirected graph $\mathcal G$ exhibits the following property:
\begin{property}\label{pro:L:eig}
For an undirected graph $\mathcal G$, the eigenvalues of its Laplacian matrix, $\lambda_i$ $(i=1,\ldots,N)$, are real and can be ordered such that $0=\lambda_1\leq \lambda_2\leq \ldots\leq \lambda_N$.
Additionally, if $\mathcal G$ is connected, then $\lambda_2$ is positive and called as the algebraic connectivity of the graph.
\end{property}

Given a graph $\mathcal G(\mathcal W)$, let us denote $\mathcal N_i$ as the set of the neighbors of the $i^{th}$ robot, which has a direct edge to the $j^{th}$ robot for $j\in \mathcal N_i$.
Thus, the $i^{th}$ robot is able to obtain the displacement to $j\in \mathcal N_i$ by using the proximity sensors on the $i^{th}$ robot.
By denoting the displacement vector $r_{ji}(t)=x_i(t)-x_j(t)$, the distance between the $i^{th}$ and $j^{th}$ robots are given as $d_{ji}(t)=\|r_{ji}(t)\|$, where $\|\cdot\|$ denotes the Euclidean norm of the enclosed vector.
Since the local displacement measurement is mutual, the sensory topology $\mathcal G(\mathcal W)$ being undirected leads to $r_{ji}(t)=-r_{ij}(t)$ and $d_{ji}(t)=d_{ij}(t)$. The formation of the mobile robot network is constructed by utilizing $r_{ji}$ according to the inter-agent measurement topologies without data exchange\footnote{The argument of time dependent signals is omitted, for example $r_{ji} \equiv r_{ji}(t)$, unless otherwise required for the sake of clarity.}
%Moreover, we let $i=\{1,\cdots, N\}$ for the $i^{th}$ agent in the mobile robot network, and $j\in\mathcal N_i$ denotes the adjacent agents in the neighborhood of the $i^{th}$ agent.}}.
Thus, we have the next assumption for the proposed networked  system:
\begin{assumption}
There is no inter-agent communication or data exchange in the robot network.
\end{assumption}

\subsection{Global and Local Formation Tracking}
The time-varying global trajectory of the desired formation with respect to $\Sigma_W$, the world coordinates, is predefined and denoted by $h^d(t):\left [ 0, \infty \right )\rightarrow \mathcal Q$, which is a twice differentiable continuous function of time.
As illustrated in Fig.~\ref{fig:illustration}, the desired formation of the mobile robots are described by $\bar h_i^d(t):\left [ 0, \infty \right )\rightarrow \mathcal Q$, which are time-varying continuous vectors, with respect to $h^d(t)$.
Subsequently, the time-varying desired trajectories for each of the mobile robots, $h_i^d(t)\in\mathcal Q$, are given as $h_i^d(t)=h^d(t)+\bar h_i^d(t)$, $i=1,\ldots,N$.
For the mobile agents to achieve time-varying formation, the desired relative displacement between the $i^{th}$ robot and its neighbors, $j\in\mathcal N_i$, are expressed by $h_{ji}^d(t)=\bar h_i^d(t)-\bar h_j^d(t)$, which further leads to $h_{ji}^d(t)=h_i^d(t)-h_j^d(t)$ and $h_{ij}^d(t)=-h_{ji}^d(t)$.

Let $\mathcal X_0$ be a subset of the compact set $\mathcal Q$ such that $\mathcal X_0\subseteq \mathcal Q$ is closed and bounded. The time-varying formation tracking can be defined for local formation tracking and global formation tracking, respectively.

%The boundedness is first defined as:
%\begin{definition} \cite{khalil} \label{def:bound}
%The solution of $\dot{x} = f(x)$ is bounded if there exists a positive constant $c$, independent of $t_0 \geq 0$, and for every $a \in \left (0, c\right )$, there is $\beta = \beta(a) > 0$, independent of $t_0$, such that 
%$\left \| x(t_0) \right \| \Rightarrow \left \| x(t) \right \| \leq \beta, \forall t \geq t_0$.
%\end{definition}
%}

\begin{definition}\label{def:local}(Local Formation Tracking)
The networked robot system~\eqref{eq:dyn:rob} is said to achieve local formation tracking if for any given initial states $x_i(0)\in\mathcal X_0$, $i=1,\ldots,N$, the states satisfy that
$
\lim_{t\rightarrow\infty}\left[(x_i(t)-x_j(t))-(\bar h_i^d(t)-\bar h_j^d(t))\right]=0,
$
for all $j\in\mathcal N_i$, which implies that
$
\lim_{t\rightarrow\infty}r_{ji}(t)=\lim_{t\rightarrow\infty}h_{ji}^d(t).
$
\end{definition}

\begin{definition}\label{def:global}(Global Formation Tracking)
The networked robot system~\eqref{eq:dyn:rob} is said to achieve global formation tracking if  for any given initial states $x_i(0)\in\mathcal X_0$, $i=1,\ldots,N$, the states satisfy that
$
\lim_{t\rightarrow\infty}\left(x_i(t)-h_i^d(t)\right)=0.
$
\end{definition}

The problem can be considered as a hierarchical framework consisting of the local and global formation tracking systems.
Thus, in the proposed mobile robot network, global formation tracking (Definition~\ref{def:global}) is the sufficient condition for local formation tracking, and the local formation tracking (Definition~\ref{def:local}) is necessary for global formation tracking.

\subsection{Problem Formulation}

The time-varying formation tracking for a mobile robot network is studied in this paper, as illustrated in Figure~\ref{fig:illustration}.
Based on the dynamics~\eqref{eq:dyn:rob}, the $N$ mobile agents are controlled to maintain a time-varying formation based on the global positioning and local relative displacement measurements.
The global positioning measurements can be obtained from the GPS (global positioning system) outdoors or other indoor/outdoor positioning and tracking systems.
It is noted that we use the term `GPS' generally to refer to any global positioning systems, e.g. location based service in mobile devices, WiFi positioning system, or optical localization system.%~\cite{Bonnifait98TRA,Ouyang10TVT,Prieto21TSP}.
In the proposed system, each robot can obtain global positioning data for the global formation tracking as addressed in Definition~\ref{def:global}.

The local measurement is implemented by extrinsic sensors on each of the mobile agents to obtain the relative displacement to all its neighbors.
Moreover, the relative velocity between two adjacent robots can also be obtained accordingly.
The local positioning information is utilized to control the robot network for achieving local formation tracking as stated in Definition~\ref{def:local}.
Since there is no communication between the mobile agents, the desired trajectories of an agent's neighbors are important in achieving the time-varying formation tracking.
Thus, we have the next assumption:
\begin{assumption}\label{assump:des:traj}
The desired formation trajectories $h^d(t)$, $h_i^d(t)$, and $h_j^d(t)$ for $j\in\mathcal{N}_i$ are available to the $i^{th}$ robot.
\end{assumption}

We consider the situation that the signals transmitted from the global positioning systems to the mobile robots may go under deception attacks\cite{Book04Mahmoud,Modares20TCYB,Hwang10TCST}.
Let us define the position of the $i^{th}$ robot from the global positioning system as $x_i^g(t)$, which might be different from the actual position $x_i(t)$ depending on condition of the positioning system.
If the global positioning signal is accurate, we have $x_i^g(t)=x_i(t)$.
However, if the global positioning signal is under deception attacks, then the erroneous position data $\hat x_i^g(t)$ is described as 
\begin{align} \label{eq:deception_attack}
    \hat x_i^g(t)=\delta_{xi}(t)x_i(t)+\Delta_{xi}(t),
\end{align}
where $\delta_{xi}(t)\in \mathbb R$ is a time-varying function, and $\Delta_{xi}(t)\in\mathbb R^n$ is a continuous function.
\section{Time-varying Formation Tracking}\label{sec:main}

\subsection{Formation Tracking Control Design and Analysis}\label{sec:form:track}
For dynamically controlled mobile agents, let us consider the time-varying formation tracking control
\begin{eqnarray}\label{eq:control:input}
&&\hspace{-.37in} u_i(t)=\ddot{h}_i^d-\sigma_{fi}(\dot x_i-\dot{h}_i^d)-\kappa_{gi}f^g_i(t)-\kappa_ff_i^l(t),
\end{eqnarray}
where $\sigma_{fi}{\in \mathbb R^{n \times n}}$ is a positive-definite matrix, $\kappa_f,\kappa_{gi} \in \mathbb R$ are positive gains for local and global formation tracking, respectively, and
\begin{eqnarray}
&&\hspace{-.45in} f_i^l(t)=\sum_{j\in\mathcal N_i}w_{ji}(\dot r_{ji}-\dot h_{ji}^d)+\sigma_{fi}\sum_{j\in\mathcal N_i}w_{ji}\left(r_{ji}-h_{ji}^d\right),\\
&&\hspace{-.45in} f_i^g(t)=(\dot x_i-\dot{h}_i^d)+\sigma_{fi}\left(x^g_i-h^d_i\right)
\end{eqnarray}
are the formation tracking control commands.
In the control input~\eqref{eq:control:input}, the relative displacement $r_{ji}$ and velocity $\dot r_{ji}$ are obtained by the sensors mounted on each of the mobile robots as mentioned in Section~\ref{sec:prob}.
The velocity of the mobile robot $\dot x_i$ with respect to $\Sigma_W$ can be obtained from inertial navigation systems, e.g. an inertial measurement unit (IMU), or an odometry system such as visual odometry.
It is noted that the global positioning data, $x_i^g(t)$, are required in formation tracking because the absolute position obtained from the integration of the velocity data are not reliable or accurate enough to be implemented in the formation control due to the accumulative drifting errors.

\begin{assumption}\label{assump:ini:pos}
All robots start from initial positions that are known or measurable by its neighbors.
\end{assumption}

By denoting $\tilde x_i:=x_i-h^d_i$ as the global tracking error and $e_i=\sum_{j\in\mathcal N_i}w_{ji}(\tilde x_i-\tilde x_j)$ as the accumulated inter-agent errors with respect to the weighted Laplacian $L(\mathcal G)$, we have $r_{ji}-h_{ji}^d=\left(x_i-x_j\right)-(h_i^d-h_j^d)={\tilde x}_i-{\tilde x}_j$.
If the position of the mobile agents in the robot network can be obtained accurately such that $x_i^g(t)=x_i(t)$, the closed-loop dynamics is given as
$
%&&\hspace{-.3in}
\ddot{\tilde x}_i+\kappa_{gi}\dot{\tilde x}_i+\kappa_f\dot e_i=-\sigma_{fi}\dot{\tilde x}_i-\sigma_{fi}\kappa_{gi}\tilde x_i-\sigma_{fi}\kappa_fe_i.
$
Given $\xi_i=\dot{\tilde x}_i+\kappa_{gi}\tilde x_i+\kappa_fe_i$, the above equation becomes
$\label{eq:cls:xi}
%&&\hspace{-.2in}
\dot \xi_i=-\sigma_{fi}\xi_i.
$
Hence, under Assumptions~\ref{assump:local:mea} through \ref{assump:ini:pos},  we have the following result for the mobile robot network with perfect global positioning information.

\begin{theorem}\label{thm:global}
For the mobile robot network described by~\eqref{eq:dyn:rob} under an undirected and connected displacement sensory graph $\mathcal G$ with weighted Laplacian matrix $L(\mathcal G)$, if the global positioning data are accurate such that $x_i^g(t)=x_i(t)$ for $i=1,\ldots, N$, then the control input~\eqref{eq:control:input} guarantees that the mobile robot network achieves global formation tracking.
\end{theorem}

A proof of Theorem~\ref{thm:global} is provided in Appendix\ref{proof:thm:global}.
This theorem demonstrates that guarantee of the global formation tracking (Definition~\ref{def:global}) can also ensure local formation tracking (Definition~\ref{def:local}) as addressed in Section~\ref{sec:prob}.
The tracking performance of the local and global formation depends heavily on the values of $\kappa_f$ and $\kappa_{gi}$ in~\eqref{eq:dxi:cls}.
Let us first consider the case where there is only global formation tracking such that $\kappa_f=0$, then
the closed-loop dynamics becomes $\ddot{\tilde x}_i+\kappa_{gi}\dot{\tilde x}_i=-\sigma_{fi}\dot{\tilde x}_i-\sigma_{fi}\kappa_{gi}\tilde x_i$.
By denoting $\xi_i^g=\dot{\tilde x}_i+\kappa_{gi}\tilde x_i$, the given closed-loop system becomes $\dot\xi_i^g=-\sigma_{fi}\xi_i^g$.
It is noted that there is no inter-agent term in the closed-loop dynamics because this case considers only global positioning signals; thus, the mobile robot network becomes a decoupled system for trajectory tracking.
Hence, by considering $V_i^g(\xi_i^g)=\frac{1}{2}(\xi_i^g)^T\sigma_{fi}^{-1}\xi_i^g$, we have $\dot V_i^g=-(\xi_i^g)^T\xi_i^g$, which is negative definite.
Therefore, we conclude that $\lim_{t\rightarrow\infty}\xi_i^g(t)=0$ by following the proof of Theorem~\ref{thm:global}.
From the closed-loop control system, we further have $\dot{\tilde x}_i=-\kappa_{gi}\tilde x_i+\xi_i^g$.
As shown in~\cite{Liu12TROSyn,sontagcics}, if $\xi_i^g(t)$ is a signal that asymptotically converges to zero, and $\tilde x_i$ is bounded, then $\lim_{t\rightarrow\infty}\tilde x_i(t)=0$, which implies that $\lim_{t\rightarrow\infty}\left(x_i(t)-h_i^d(t)\right)=0$, $i=1,\ldots, N$.
Consequently, the mobile robot network achieves global formation tracking as stated in Definition~\ref{def:global}.

Next, let us consider the case of formation tracking with only local measurements such that $\kappa_{gi}=0$.
For the mobile robot network with an undirected and connected displacement sensory graph $\mathcal G$, the closed-loop dynamics is described as $\ddot{\tilde x}_i+\kappa_f\dot e_i=-\sigma_{fi}\dot{\tilde x}_i-\sigma_{fi}\kappa_fe_i$.
From $\xi_i^l=\dot{\tilde x}_i+\kappa_fe_i$, the stacked dynamics $\dot\xi_i^l=-\sigma_{fi}\xi_i^l$ can be proved to be asymptotically stable by considering $V^l(\xi_i^l)=\frac{1}{2}(\xi^l)^T\sigma_{fi}^{-1}\xi^l$, where $\xi^l$ is the stacked vector of $\xi_i^l$.
With $\dot V^l=-(\xi^l)^T\xi^l$, we obtain that $\xi^l\in\mathcal L_2\cap\mathcal L_{\infty}$ and $\lim_{t\rightarrow\infty}\xi^l(t)=0$.
By following~\eqref{eq:dxi:cls} in the proof of Theorem~\ref{thm:global}, the stacked form is given as $\dot{\tilde x}=-[(\kappa_fL)\otimes I_n]\tilde x+\xi^l$ with $\kappa_{gi}=0$.
For $\xi^l(t)$ converging to the origin, the mobile robot network achieves consensus to the agreement set that is the subspace of $span\{1\}$ such that $\tilde x_i(t)=\tilde x_j(t)$, $i=1,\ldots, N$, $j\in\mathcal N_i$ when $t\rightarrow\infty$~\cite{Book10Egerstedt}.
Therefore, from the definition of $\tilde x_i(t)$, we obtain that $\lim_{t\rightarrow\infty}r_{ji}(t)=\lim_{t\rightarrow\infty}h_{ji}^d(t)$ so that the local formation tracking is guaranteed.

\begin{comment}
\begin{remark}
According to the above discussion, the relationship between the global and local formation tracking to the control gains $\kappa_{gi}$ and $\kappa_f$ is significant.
The priority of global formation tracking can be ensured from the control term $f_i^g(t)$, and the local control term $f_i^l(t)$ can guarantee the inter-agent displacement to maintain a time-varying formation.
It is noted that the proposed control system does not require global localization of a robot for local maneuver.
{\it Therefore, the system can be considered as redundant for local formation tracking such that when the mobile agents lose global positioning signals, the mobile robot network can still achieve time-varying local formation.}
\end{remark}
\end{comment}

\subsection{Malicious Attacks on Global Positioning Signal}\label{sec:main:attack}

%探討deception attack的參數對系統之影響
%定義不同參數組合之攻擊名稱

By following the analysis in Section~\ref{sec:form:track}, we obtain that the local formation tracking can be ensured with or without global positioning information.
%Thus, if the global positioning signals are under malicious attack, the mobile robot network can also ensure local formation with the sacrifice of global formation tracking.
%Various types of attacks on the global positioning data are studied in this section, and their effects on the stability and tracking performance are addressed.
In the proposed system, the resilience of the networked robot system is guaranteed based on the inter-agent position information.
If the formation tracking errors are abnormally large due to initial conditions, erroneous judgement could occur and significantly influence the resilient efficacy.
Therefore, in this system, we consider the following assumption:

\begin{assumption}\label{assum:attack:time}
The mobile robot network has achieved asymptotic stability before any agents are subject to adversarial attacks on global positioning signals such that attacks occur at $t=t_{ai}\geq 0$ and $q_i\in \mathcal Q_{as}$, $\forall i=1,\ldots,N$, where $\mathcal Q_{as}=\{q_i:=(\tilde x_i^T, \dot{\tilde x}_i^T)^T~|~\tilde x_i=\dot{\tilde x}_i=0\}$.
\end{assumption}

The influence on the system due to malicious attack depends on its features~\cite{Hwang10TCST}. 
For a deep understanding of the deception attack defined in \eqref{eq:deception_attack}, it is classified into three types, named additive attack, hybrid attack, and unstable attack, to investigate the effect of each parameter in \eqref{eq:deception_attack}.
The deception attack with $\delta_{xi} = 1$ is an additive attack; otherwise it's a hybrid attack.
Compared with additive attack, hybrid attack is more related to original data $x_i$.
Additionally, unstable attack is defined as $\hat x_i^g(t)=x_i(t)-c_{ai}\xi_i(t),$ where $c_{ai}$ is a coefficient of the attack model.
By choosing the same Lyapunov function candidate in Theorem~\ref{thm:global}, it is derived as $\dot V_i=(\kappa_{gi}c_{ai}-1)\xi_i^T\xi_i$. Obviously, if $\kappa_{gi}c_{ai}>1$, $\dot V_i$ is positive definite, which means the system is unstable.

\subsection{Performance Index}\label{sec:main:index}

%In the time-varying formation tracking system for the mobile robot network, the local/global tracking gains, desired trajectory, and types of attacks would significantly influence the performance of the entire system.
%Moreover, to quantitatively assess the resilience of the mobile robot network, an appropriately defined performance index is extremely crucial.
In this section, we design a global performance index to quantify the tracking performance of the mobile robot network and evaluate the system's performance under various global positioning attacks through simulation.
It is noted that, to the best of the authors' knowledge, this is the first performance index proposed to mobile robot network on time-varying formation tracking.

The design of the performance index depends on both local and global formation tracking.
The global formation tracking errors for the $i^{th}$ agent has been defined previously as $\tilde x_i=x_i-h_i^d$.
For local tracking, we define $e_i^1=\sum_{j\in \mathcal N_i}(\tilde x_i-\tilde x_j)$ as the summation of local formation tracking error with a unity weight\footnote{The design of $e_i^1$ is similar to the inter-agent errors $e_i$ defined in Section~\ref{sec:main} and multi-agent consensus, but this term, without the weights for the graph Laplacian, is considered in the evaluation of the local tracking performance.}.
Furthermore, the average of local formation errors is given as $\bar e_i^1=e_i^1/n_i$, where $n_i$ is the number of the neighbors of the $i^{th}$ mobile agent.
Hence, the performance index is defined as
\begin{eqnarray}\label{eq:perf:index}
&&\hspace{-.5in} \mathcal I_f^g=\left(\vartheta+\sum_{i=1}^N \|\bar e_i^1\|\right)\!/\!\left(\vartheta+\sum_{i=1}^N \|\bar e_i^1\|+\alpha_f^g\sum_{i=1}^N \|\tilde x_i\|\right)\!\!,
\end{eqnarray}
where $\vartheta$ and $\alpha_f^g$ are positive constants.
It is noted that $\mathcal I_f^g$ is defined globally for the formation tracking control.

Based on $\mathcal I_f^g$, if the mobile robot network achieves global formation tracking, then the local formation tracking is also guaranteed; therefore, $\tilde x_i$ and $\bar e_i^1$ all converging to zero gives that $\mathcal I_f^g=1$.
If the mobile robots only achieve local formation tracking but fail to track the global formation, then $\bar e_i^1$ converges to zero but $\tilde x_i$ could be non-zero.
Thus, $\mathcal I_f^g$ is less than one, and its actual value depends on the tracking errors of the global formation.
If the system becomes unstable, then $\tilde x_i$ diverges so that $\mathcal I_f^g$ goes to zero. 
Hence, the index $\mathcal I_f^g$ tells the possibility of the mobile robots being under deception attack on the global positioning data.

\begin{comment}
%The performance index is utilized to evaluate the global and local formation tracking system under various attacks.
When the mobile robot network achieves asymptotic stability (Theorem~\ref{thm:global}), we can show that $\mathcal I_f^g$ goes to one.
Subsequently, if some of the mobile agents in the network is under attack, then $\tilde x_i$ and $\bar e_i^1$ are both non-zero so that $\mathcal I_f^g$ is less than one.
Therefore, the performance index is designed to describe the possibility of the mobile robots being under attack on the global positioning data under the problem formulation.
Additionally, the proposed resilience index will be utilized in our experiments to evaluate the performance of position recovery and resilient operation presented in Section~\ref{sec:CoL}.
\end{comment}

\begin{figure}
   \centering
\hspace{-.1in}   \includegraphics[width=1.4in]{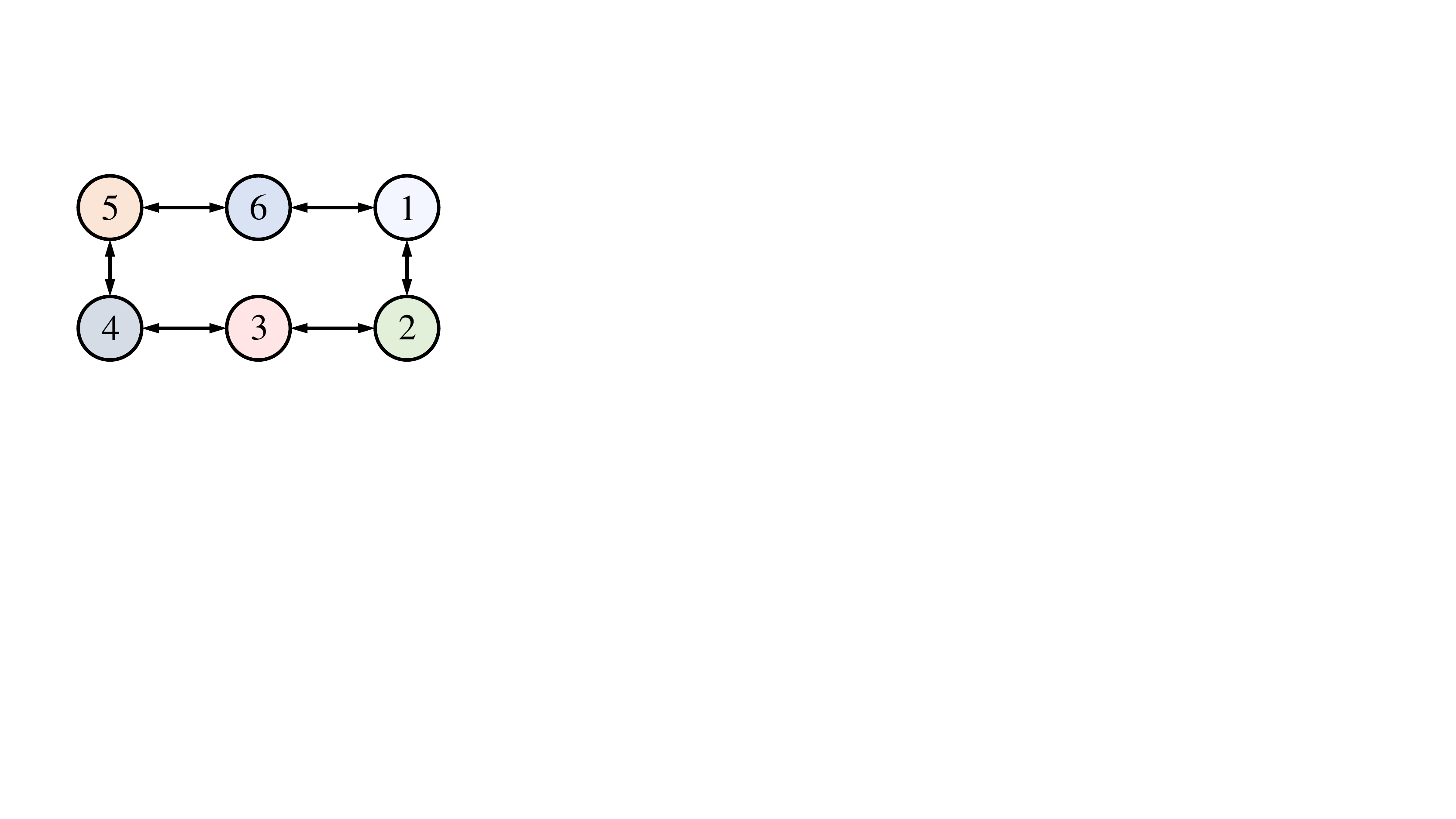}
   \caption{Sensory topology of the mobile agents in the networked robotic system considered in the simulation analysis.}
\label{fig:sim:topology}
\end{figure}

\subsection{Numerical Examples - Formation Tracking and Attacks}\label{sec:main:sim}
\begin{comment}
The functionality of the unaffected systems is significant to the evaluation of the resilience and efficiency of a networked robotic system.
Especially for the research topic in this paper, the differences in the number of mobile agents, time-varying formation, and desired trajectory all make the evaluation of networked system performance a critical issue.
The performance index, presented in the previous section, is validated via simulation to show the adverse effects on the mobile robot network in time-varying formation tracking subjected to the aforementioned malicious attacks.
\end{comment}

We consider a mobile robot network composed of six agents under the sensory topology illustrated in Fig.~\ref{fig:sim:topology}.
The control algorithm~\eqref{eq:control:input} is utilized to control the mobile robot network to form a hexagon, rectangle, and triangle at $t=0\sim15$ sec, $t=15\sim30$ sec, and after $t=45$ sec, respectively.
Moreover, the formation is considered to track a lemniscate time-varying trajectory described by
$h^d(t)=\big[-5\sin(2\pi t/15)/\big(2(\cos(2\pi t/15)-3)\big), -12\cos(\pi t/15)/\big(5(\cos(2\pi t/15)-3)\big)\big]^T$.
The control gains are selected as $\kappa_f=2$, $\sigma_{fi}=I_2$, $\kappa_{gi}=2$ with an identical sensory weights of $w_{ji}=1$, where $I_n\in\mathbb R^{n\times n}$ denotes an identity matrix.
The performance index~\eqref{eq:perf:index} is utilized to evaluate the formation tracking in the following simulation with $\vartheta=10$ and $\alpha_f^g=5$.

\begin{comment}
\begin{figure}
   \centering
\hspace{-.1in}   \includegraphics[width=3.35in]{Figure/M_traj_noAttack}
   \caption{Performance index ($\mathcal I_f^g$) and trajectory snapshots of the mobile robot network without global positioning attack. The vertical dashed lines in the sub-figure for $\mathcal I_f^g$ represent the time of snapshots in the other three sub-figures.}
\label{fig:sim:noattack}
\end{figure}
\end{comment}

\begin{figure}
   \centering
\hspace{-.1in}   \subfloat[\label{fig:sim:noattack} ]{\includegraphics[width=3.35in]{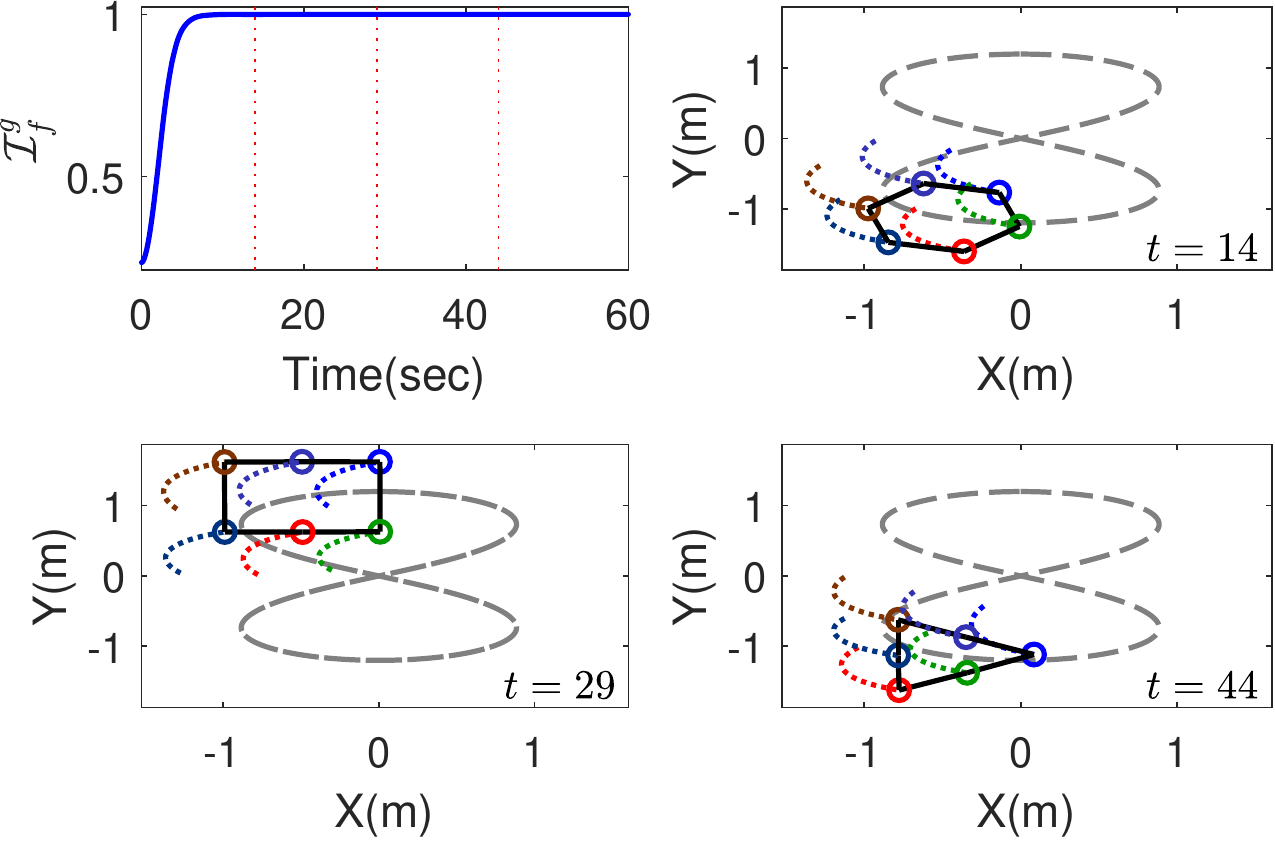}} \\
\hspace{-.1in}   \subfloat[\label{fig:sim:mixedattack}
]{\includegraphics[width=3.35in]{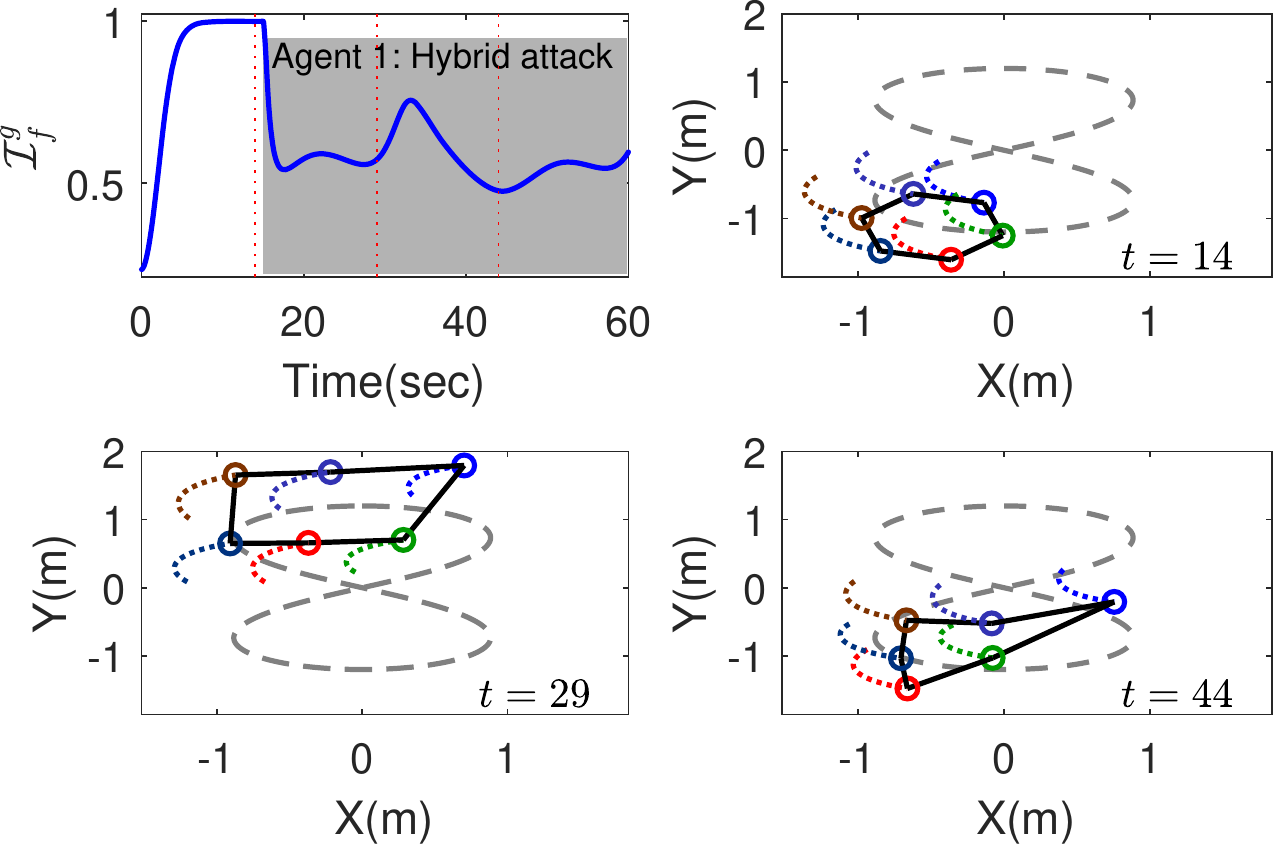}}
   \caption{Performance index ($\mathcal I_f^g$) and trajectory snapshots of the mobile robot network (a) without (b) with global positioning attack. The vertical dashed lines in the sub-figure for $\mathcal I_f^g$ represent the time of snapshots in the other three sub-figures.}
%\label{fig:sim:noattack}
\end{figure}

For the mobile robot network that is not under positioning attack, the agents' trajectories and performance index are shown in Fig.~\ref{fig:sim:noattack}.
With the unaffected global positioning signals and local displacement measurement, the mobile agents are able to stably keep a time-varying formation while tracking a trajectory.
The performance index converges to one if the system is asymptotically stable with guaranteed local and global formation tracking.
Next, we consider the case where Agent 1 is under hybrid attack ($\delta_{x1}=2$ and $\Delta_{x1}=[-2, -2]^T$) starting at $t_{a1}=15$ sec, which is after the networked robotic system achieving asymptotic stability as stated in Assumption~\ref{assum:attack:time}.
The results in Fig.~\ref{fig:sim:mixedattack} show that the hybrid attack affected the motion of Agent 1, which is marked in blue, and the adjacent agents of Agent 1 also deviated from their positions in the desired local formation.
Due to malicious positioning attack, the performance index decreased far from one after $t=t_{a1}$, the starting time of the attacks.

\begin{comment}
\begin{figure}
   \centering
\hspace{-.1in}   \includegraphics[width=3.35in]{Figure/M_traj_hybrid}
   \caption{Performance index and trajectory snapshots of mobile robot network with M-hybrid attack on Agent 1 starting at $t_{a1}=15$ sec. The grey area represents the duration of the mobile agent under M-hybrid attack with $\delta_{x1}=2$ and $\Delta_{x1}=[-2, -2]^T$.}
\label{fig:sim:mixedattack}
\end{figure}
\end{comment}

\begin{figure}
   \centering
\hspace{-.1in}   \includegraphics[width=3.35in]{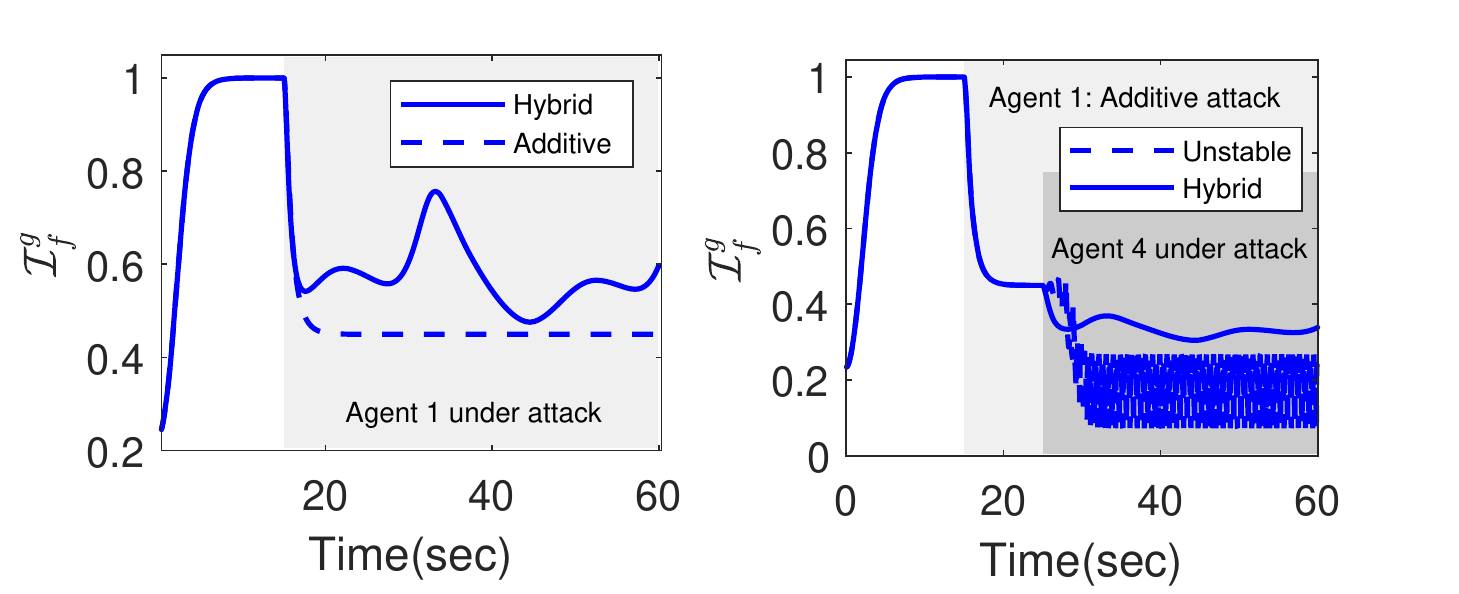}
   \caption{Performance indices of mobile robot network subjected to different malicious attacks on global positioning. Left: Only Agent 1 is under attack at $t_{a1}=15$ sec under either an hybrid attack (as in Fig.~\ref{fig:sim:mixedattack}) or additive attack. Right: Agent 1 is under additive attack at $t_{a1}=15$ sec, and subsequently Agent 4 is under either hybrid or unstable attacks at $t_{a4}=25$ sec. ($\delta_{x1}=\delta_{x4}=2$, $\Delta_{x1}=\Delta_{x4}=[-2, -2]^T$, and $c_{a4}=5$)}
\label{fig:sim:attack:comp}
\end{figure}

%Figs.~\ref{fig:sim:noattack} and~\ref{fig:sim:mixedattack} demonstrate the performance of the proposed time-varying formation tracking control in mobile robot network with and without adversarial attacks.
To compare the evolution of the performance index with respect to different types of attacks designed in Section~\ref{sec:main:attack}, we have conducted various analyses shown in Fig.~\ref{fig:sim:attack:comp}.
If the additive attack is with a constant bias signal, then the performance index decreases to a smaller value without time-dependent variation; however, under the hybrid attack, resulting from the multiplicative type $\delta_{xi}$, the variation of the performance index is dependent on the formation and desired trajectory.
From the case where both Agent 1 and Agent 4 were under attack, we can observed that multi-agent attacks would degrade the performance index.
Moreover, the unstable attack, due to the unstable term $-c_{ai}\xi_i$, would cause oscillations in the performance index, which results from the constant switching between the local and global formation tracking control.

%>>>>>>>>>>>>>>>>>>>>>>>>>>>>>>>>>>>>>>>>>>>>>>>>>>>>>>>>>>>>>
\begin{comment}
In addition to quantifying the effects of attacks, the performance index can also be utilized to assess the converging rate of the formation tracking control.
As stated in Section~\ref{sec:form:track}, the proposed time-varying formation tracking control is a redundant system since all mobile agents have global positioning information.
Therefore, if a portion of the mobile agents lose access to the global positioning system, the formation tracking can still be guaranteed through the use of local formation control.
However, the convergence rate would be effected dramatically.
Fig.~\ref{fig:sim:kappag:comp:1} illustrates the performance index, in the absence of attack, for the proposed TVFC with respect to different numbers of agents losing global positioning, denoted as $N_{g0}$.
The convergence rate decreases along with the increase of $N_{g0}$, and when all mobile agents lose global positioning signals, the mobile robot network can only achieve local formation with a relatively low performance index.
\end{comment}
%<<<<<<<<<<<<<<<<<<<<<<<<<<<<<<<<<<<<<<<<<<<<<<<<<<<<<<<<<<<<<<<<<<<

%%%%%%%%%%%%%%%%%%%%%%%%%%%%%%%%%%%%%%%%%%%%%%%%%%%%%%%%%%%%
\section{Cooperative Localization and Resilient Operation}~\label{sec:CoL}
%Conventionally, accurate localization is often assumed when studying the control of multi-agent systems.
%In this work, the control and localization of the robot network is co-investigated to achieve practical resilience. 
The main focus here is to improve the localization resilience toward potentially faulty global positioning information due to system-originated errors or malicious attacks.
For the rest of the paper, we generally refer to the external positioning system that is subject to attacks as GPS.
By default, the global tracking of the robot network relies on the localization information from the GPS.
Since inter-agent observations are available, we introduce a cooperative localization scheme as the secondary localization approach in the case of GPS failures or attacks. In this section, we first present a communication-free cooperative localization scheme based on the EIF and then discuss the self-identification and exclusion of GPS attacks.

\begin{comment}
\begin{figure}
   \centering
\hspace{-.1in}   \includegraphics[width=3.35in]{Figure/M_N_aComparison}
   \caption{Performance indices with respect to different numbers of mobile agents without global positioning in the absence of malicious attacks. The number of mobile agents with $\kappa_{gi}=0$ is denoted by $N_{g0}$.}
\label{fig:sim:kappag:comp:1}
\end{figure}
\end{comment}

\subsection{Positioning Recovery}\label{sec:CoL:recovery}
Localization resilience is achieved through redundancy, i.e., a sensor fusion mechanism consisting of proprioceptive sensor measurements, GPS measurements, and inter-agent relative measurements with respect to neighboring agents. It is assumed that only the GPS measurements may be subject to attacks.
The EIF was chosen as the information fusion method in favor of its efficiency in incorporating multi-modal sensor measurements in multi-step filter updates. Here we only include the details necessary to understand the propose positioning recovery method. For a comprehensive treatment of EIF, the readers are referred to \cite{Jazwinski:70a}.

For applications in a three-dimensional space, the state vector for the $i^{th}$ agent at time $k$ consists of the location of the agent defined as $x_i(k) := [x(k); y(k); z(k)]^T \in \mathbb{R}^3$.
The agent's motion model can be found as

\begin{eqnarray}\label{eq:motion_model}
x_i(k) = x_i(k-1) + [\dot{x}_i(k) + \nu_i(k)]\Delta t,
\end{eqnarray}
where $\dot{x}_i(k) \in \mathbb{R}^3$ denotes the agent velocity that can be measured directly, and $\nu_i(k) \sim \mathcal{N}\big(0, Q(k)\big)$ is the random process noise that follows a zero-mean Gaussian distribution with covariance matrix $Q(k) \in \mathbb{R}^{3\times 3}$. The state vector prediction at time $k$, $\hat{x}_i(k|k-1)$, can be calculated as
\begin{equation}\label{eq:state_estimate}
    \hat{x}_i(k|k-1) = \hat{x}_i(k-1|k-1) + \dot{x}_i(k) \Delta t.
\end{equation}
The covariance of the location estimate for the $i^{th}$ agent, $P_i \in \mathbb{R}^{3 \times 3}$, is propagated accordingly as

\begin{align}
    P_{i}(k|k-1) = F_i(k)P_i(k-1|&k-1)F_i(k)^T \nonumber \\
    & + G_i(k)Q(k)G_i(k)^T, \label{eq:P_predict}
\end{align}
where $F_i(k)$ and $G_i(k)$ are the Jacobian matrices calculated based on the state estimate from \eqref{eq:state_estimate}. Here, $F_i(k) = I_{3}$ and $G_i(k) = \Delta t \cdot I_{3}$. The information matrix, $\Phi_i \in \mathbb{R}^{3\times 3}$, and information vector, $\varphi_i \in \mathbb{R}^3$, can be found as $\Phi_i(k|k-1) = P_i(k|k-1)^{-1}$ and $\varphi_i(k|k-1) = \Phi_i(k|k-1)\,\hat{x}_i(k|k-1)$, respectively.

During normal operations, each agent updates its location estimate with the GPS measurement. In the event that the GPS measurement is deemed unreliable or faulty, the agent performs location update through cooperative localization instead.
The measurement models for these two types of updates are

\begin{align}
    s^\text{GPS}_i(k) &= x_i(k) + \omega^\text{GPS}_i(k), \label{eq:update_gps}\\
    s^{r_{ij}}_i(k) &=  x_j(k) - x_i(k) + \omega^{r_{ij}}_i(k), \label{eq:update_r}
\end{align}
where $\omega^\text{GPS}_i(k)$ and $\omega^{r_{ij}}_i(k)$ are the random noise signals associated with the GPS or relative position measurements, both of which are assumed to follow zero-mean Gaussian distributions with covariance matrices $R^\text{GPS}(k)$ and $R^{r_{ij}}(k)$, respectively, and $j\in\mathcal N_i$ denotes the index of the neighboring agents.
Note that \eqref{eq:update_r} does not require information to be sent from the neighbors because the neighbors' desired trajectories are known to others.  For either update mode, denoted by a superscript ${\square} \in \{\text{GPS},\,r_{ij} \}$, the updates of the information matrix and vector follow the same EIF procedure such that
\begin{align}
    \Phi_i(k|k) &= \Phi_i(k|k-1) + H^{\square}_i(k)^T R^{\square}_i(k)^{-1}H^{\square}_i(k), \label{eq:Phi_update}\\
    \varphi_i(k|k) & = \varphi_i(k|k-1) + H^{\square}_i(k)^TR^{\square}_i(k)^{-1}[s_i^{\square}(k) - \hat{s}_i^{\square}(k)  \nonumber \\
    &\hspace{.15in} + H^{\square}_i(k)\hat{x}_i(k|k-1)], \label{eq:infovec_update}
\end{align}
where $H_i^{\square}$ is the Jacobian matrix calculated from either \eqref{eq:update_gps} or \eqref{eq:update_r} with $R^{\square}_i(k)$ being the covariance matrix of the corresponding measurement.
Here, $H^\text{GPS}_i = I_{3}$ and $H^{r_{ij}}_i = -I_{3}$. The measurement predication, $\hat{s}_i^{\square}(k)$, can be found based on either of the measurement models \eqref{eq:update_gps} and \eqref{eq:update_r} as
\begin{align}
    \hat{s}^{\text{GPS}}_i(k) &= \hat{x}_i(k|k-1),\\    
    \hat{s}^{r_{ij}}_i(k) &= h^d_j(k) - \hat{x}_i(k|k-1).    
\end{align}
Using the updated information matrix and vector, the state estimation vector and covariance matrix can be recovered as
\begin{align}
    \hat{x}_i(k|k) &= \Phi_i(k|k)^{-1}\varphi_i(k|k), \label{eq:state_est_update} \\
    P_i(k|k) &= \Phi_i(k|k)^{-1}. \label{eq:cov_update}
\end{align}

\vspace{-.2in}
\begin{theorem}\label{thm:estimator_consistency}
    If the control input~\eqref{eq:control:input} guarantees that the mobile robot network achieves global formation tracking as stated in {Definition~\ref{def:global}}, the expectation of the position estimation errors, $\tilde{x}_i(k|k) := x_i(k) - \hat{x}_i(k|k)$,  $i = 1, \cdots, N$, strictly decreases after the update with inter-agent measurements using measurement model \eqref{eq:update_r}.
\end{theorem}

A proof of Theorem \ref{thm:estimator_consistency} is provided in the Appendix\ref{proof:thm:estimator_consistency}. This theorem provides a theoretical guarantee that, when global tracking is achieved asymptotically, the positioning error will decrease through inter-agent measurement updates using only the desired positions of the neighbors. When the robot network remains connected and as least one of the agent tracks its desired trajectory faithfully, converging localization performance across the entire agent network can be achieved.

\subsection{Attack Detection and Isolation}\label{sec:CoL:DI}
With the communication-free cooperative localization as the secondary positioning approach in the event of GPS attacks, proper detection and isolation of attacks are necessary.
Several fault detection and isolation solutions have been proposed~\cite{DingSX:2008a}.
We favor a residual-based method that utilizes the results of EIF-based sensor fusion as introduced previously. More specifically, the detection of the GPS attack is achieved by comparing the residual at the EIF update step with new GPS measurements.
This is realized through comparing the Kullback-Leibler (KL) divergence between the state estimate distributions before and after the update against a predefined threshold.

The KL divergence is a non-symmetric measure that quantifies the distance between two probability distributions. For two probability density functions, $p(x)$ and $q(x)$, defined on the same probability space, the KL divergence is defined as
\begin{equation}
    D_{KL} (p \,\|\, q) = \int^{+\infty}_{-\infty} p(x) \log\left(\dfrac{p(x)}{q(x)}\right)\text{d}x.
\end{equation}
In the case where both $p$ and $q$ are multivariate Gaussian with means $\mu_p$, $\mu_q$ and covariance matrices $\Sigma_p$, $\Sigma_q$, respectively, the KL divergence can be found as
\begin{align}
  \hspace{-.08in}  D_{KL} (\mathcal{N}_p \,\|\, \mathcal{N}_q) &= \dfrac{1}{2}\Bigg[
        (\mu_q - \mu_p)^T\Sigma^{-1}_q(\mu_q - \mu_p) \nonumber \\
        &\hspace{.1in}  \; + \text{tr}(\Sigma_q^{-1}\Sigma_p) - n + \ln\left(\dfrac{\det(\Sigma_q)}{\det(\Sigma_p)}\right)
    \Bigg],
\end{align}
where $n$ is the dimension of $x$.

Before each GPS measurement is used in the update step, the KL divergence between the following two probability density functions is calculated:
\begin{align}
    p_1 &= p\big(\hat{x}_i(k|k-1) \,|\, \hat{x}_i(k-1|k-1), \dot{x}_i(k)\big), \\
    p_2 &= p\big(\hat{x}_i(k|k) \,|\, s^\text{GPS}_i(k), \dot{x}_i(k)\big).
\end{align}
Since both probability density functions are Gaussian distributions, the KL divergence at time $k$ can be calculated as
\begin{align}
    &\hspace{-.05in} D^k_{KL}(p_1 \,\|\, p_2)\label{eq:KLD} \\
    &\hspace{-.05in} = \dfrac{1}{2}\Bigg\{
        \big[\hat{x}(k|k) - \hat{x}(k|k-1)\big]\Phi(k|k)\big[\hat{x}(k|k) - \hat{x}(k|k-1)\big]^T  \nonumber\\
        &\hspace{-.05in} + \text{tr}\left(\Phi(k|k) \cdot \Phi_i(k|k-1)^{-1}\right) + \ln\left(\dfrac{\det\left(\Phi(k|k-1)\right)}{\det\left(\Phi(k|k)\right)}\right) - n
    \Bigg\}. \nonumber
\end{align}
A GPS measurement is detected as a faulty signal when the value of $D^k_{KL}(p_1 \,\|\, p_2)$ is above a predefined threshold $\chi$. In addition, a quality measure is defined for the global positioning information to inform the formation control:
\begin{equation}\label{eq:beta}
    \beta_i(k) = 1 - \text{sat}\left(\dfrac{D^k_{KL}(p_1 \,\|\, p_2)}{\chi}\right),
\end{equation}
where $\text{sat}: \mathbb{R} \rightarrow \mathbb{R}$ is the standard saturation function defined as $\text{sat}(x) := \text{sign}(x)\min(1, |x|)$.

Algorithm~\ref{alg:1} summarizes the resilient state estimator with GPS attack detection and isolation using the criterion based on the KL divergence.
By default, the agents prioritize the global positioning information for localization. In the event of a global positioning failure or attack, the resilient state estimator switches to cooperative localization.
This switch occurs instantaneously such that the positioning of the $i^{th}$ agent is not affected by the failure of the global positioning system even if one or multiple neighbors are experiencing similar failures as long as their global tracking is faithful.

\begin{algorithm}
\caption{Resilient State Estimator for the $i^{th}$ Agent}
\label{alg:1}
\begin{algorithmic}[1]
\Require{$\hat{x}_i(k-1|k-1)$}, $P_i(k-1\,|\,k-1)$, $\dot{x}_i(k-1)$, $s^\text{GPS}_i(k)$, $\{s^{r_{ij}}_i(k) \,|\, j\in \mathcal N_i\}$, $\chi$
% \Ensure{$\hat{x}^+_i(k)$, $P_i(k\,|\,k)$, $\beta(k)$}

% \Function{Initialization}{$\boldsymbol{x}_0$, $\boldsymbol{v}_0$, $\psi_0$}
	\State Predict $\hat{x}_i(k|k-1)$ and $P_i(k\,|\,k-1)$ \Comment{\eqref{eq:motion_model}--\eqref{eq:P_predict}}
	\State Try $\hat{x}_i(k|k)$ and $P_i(k\,|\,k)$ with $s^\text{GPS}_i(k)$ \newline \phantom-  \Comment{\eqref{eq:update_gps}, \eqref{eq:Phi_update}--\eqref{eq:infovec_update}, \eqref{eq:cov_update}--\eqref{eq:state_est_update}}
	\State Compute KL divergence $D^{i,k}_{DL}$ \Comment{\eqref{eq:KLD}}
	\State Compute $\beta_i(k)$ \Comment{\eqref{eq:beta}}
	\If{$D^{i,k}_{KL} < \chi$}
	    \State Accept $\hat{x}_i(k|k)$ and $P_i(k\,|\,k)$	
	\Else
	    \State Reject $\hat{x}_i(k|k)$ and $P_i(k\,|\,k)$
	    \For{$j$ in $\mathcal N_i$} 
	        \State Compute $\hat{x}_i(k|k)$ and $P_i(k\,|\,k)$ with $s^{r_{ij}}_i(k)$
	        \newline \phantom- 
	        \Comment{\eqref{eq:update_r}, \eqref{eq:Phi_update}--\eqref{eq:infovec_update}}
	    \EndFor
	\EndIf
	\State \textbf{Return:} $\hat{x}_i(k|k)$, $P_i(k\,|\,k)$, $\beta_i(k)$
% \EndFunction
\end{algorithmic}
\end{algorithm}

% -------
\subsection{Gain-Tuning Resilient Operation}\label{sec:CoL:RO}
In addition to positioning recovery from cooperative localization, the mobile robot network in time-varying formation tracking is a redundant system with both the local and global formation tracking.
As mentioned in Section~\ref{sec:form:track}, if there is no global tracking for a portion of the mobile agents, the formation tracking can still be guaranteed.
Therefore, in this section, we propose an autonomous gain-tuning technique to strengthen the resilience of the mobile robot network by taking the advantage of redundancy.

For the proposed networked robot system under global positioning attacks, the analyses in Section~\ref{sec:main:attack} demonstrate that the tracking gain of global formation $\kappa_{gi}$ plays an important role in tracking stability.
%If the global gains $\kappa_{gi}$ of the compromised agents are set to zero, then the time-varying formation tracking can still be guaranteed with the help of local formation control as analyzed in Section~\ref{sec:form:track}.
%Especially for the three kinds of attacks discussed earlier, setting $\kappa_{gi}=0$ would lead to asymptotically stable system as stated in Assumption~\ref{assum:attack:time}.
%Additionally, we can observe from the unstable attack that if $\kappa_{gi}$ is very small, then the networked system would keep stable with larger attack gains $c_{ai}$.
Even though the networked system is under unstable attack, it would keep stable with larger attack gains $c_{ai}$ wit very small $\kappa_{gi}$.
Therefore, the manipulation of $\kappa_{gi}$ with respect to the attack identification is a useful resilient feature for the system.

The gain tuning can be designed as a function of $\beta_i\in[0,1]$, the quality measure of the global positioning signals defined in Section~\ref{sec:CoL:DI}.
If $\beta_i=1$, there is a higher confidence on the quality of the global positioning signals; whereas for $\beta_i=0$, the $i^{th}$ agent is considered under attack while the value of $D^k_{KL}(p_1 \,\|\, p_2)$ is above the threshold $\chi$.
Therefore, we can set
\begin{eqnarray}\label{eq:CL:gain:regu}
&&\hspace{-.2in} \dot \kappa_{gi}=\gamma_i\tanh\big(\sigma_{\beta_i}(\beta_i-\chi_{\beta_i})\big).
\end{eqnarray}
where $\chi_{\beta_i}\in(0,1]$ is the triggering threshold and $\sigma_{\beta_i}$ is the tuning gain for $\gamma_i$.
It is noted that $\beta_i$ encodes the confidence in the global positioning signals.
This design of resilience operation utilizes the GPS quality measure $\beta_i$ from cooperative localization, and therefore is called CL-based (cooperative-localization-based) gain-tuning approach.

\subsection{Numerical Examples - Recovery and Resilient Operation}\label{sec:resilience:sim}

\begin{figure}
   \centering
\hspace{-.1in}   \includegraphics[width=3.25in]{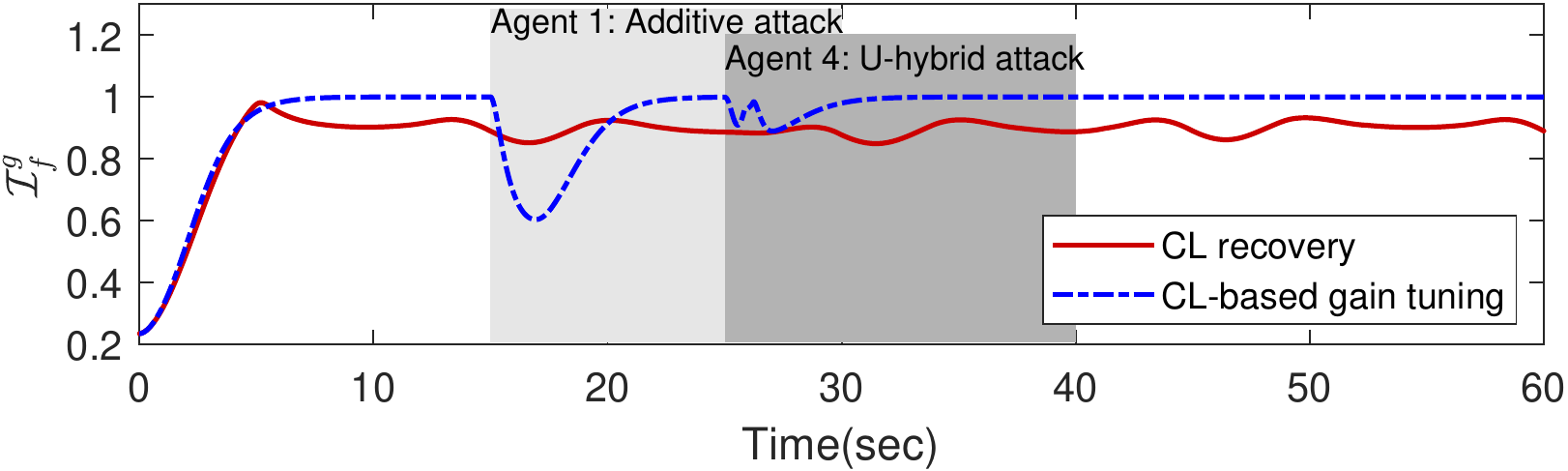}
   \caption{Performance indices under global positioning attacks with the proposed resilience operations.
   Agent 1 is under additive attack at $t_{a1}=15\sim30$ sec, and Agent 4 is under U-hybrid attack at $t_{a4}=25\sim40$ sec with $\delta_{x1}=\delta_{x4}=2$, $\Delta_{x1}=\Delta_{x4}=[-2, -2]^T$, and $c_{a4}=5$.}
\label{fig:sim:resi:index}
\end{figure}

\begin{figure}
   \centering
\hspace{-.1in}   \includegraphics[width=3.35in]{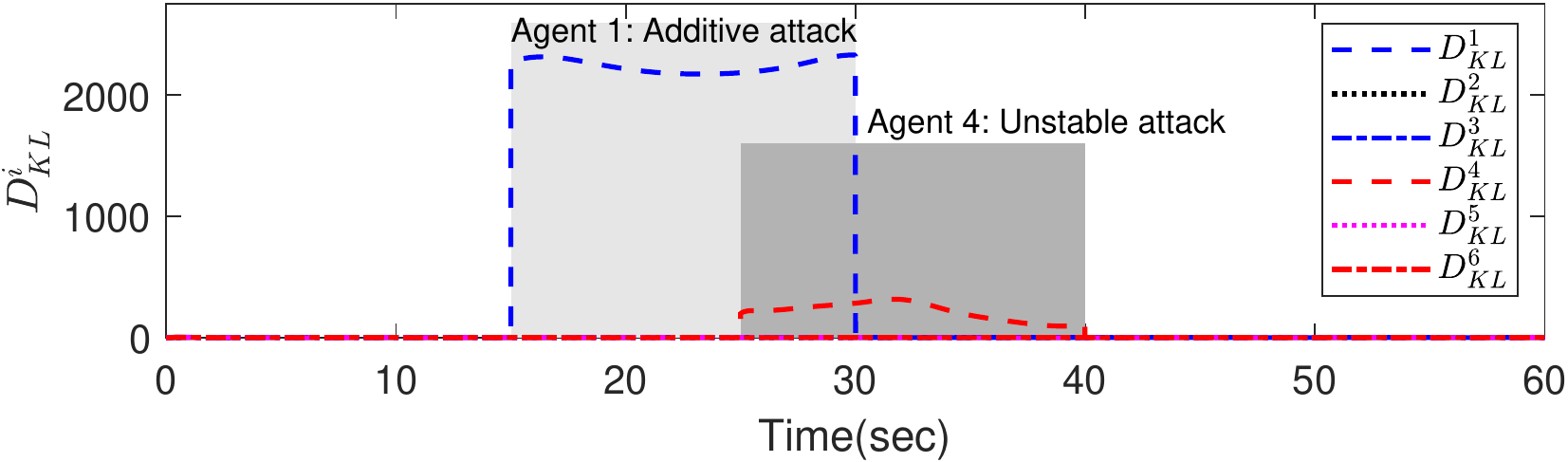}
   \caption{KL divergence of the mobile agents under global positioning attacks with  the threshold $\chi=5$.}
\label{fig:sim:resi:KLD}
\end{figure}

The effect of the proposed resilient operations is verified through numerical simulations. 
The sensory topology and control gains in the following simulation are identical to Section~\ref{sec:main:sim}.
The attack scenarios are satisfied of Assumption \ref{assum:attack:time} and similar to the case with multiple attacks in Fig.~\ref{fig:sim:attack:comp}, including the unstable attack, but both attacks are removed after 15 sec to show the resilience and recovery during and after the attacks.
%Moreover, the attacks occur after the mobile robot control system achieves asymptotic stability such that the performance index is one before the attacks.
It is noted that the resilient state estimator is active from the start of the simulation, and CL-based gain-tuning starts to regulate the global tracking gains $\kappa_{gi}$ at $t=10$ sec.
The performance indices of the resilient operations are illustrated in Fig.~\ref{fig:sim:resi:index}.

\subsubsection{CL Recovery Approach}
The cooperative localization and resilient state estimator are implemented in this section with $\chi=5$ to recover global positioning.
From Fig.~\ref{fig:sim:resi:index}, it can be seen that the CL recovery approach can maintain the performance index on a constantly high level during the positioning attacks.
%Even during the attacks (denoted by the gray areas), there is no significant degradation on the performance index using CL recovery, so the invulnerability of mobile robot network with resilient state estimator is evident.
It shows the invulnerability of mobile robot network with resilient state estimator is evident.
The evolution of KL divergence is shown in Fig.~\ref{fig:sim:resi:KLD}, where $D_{KL}^1$ and $D_{KL}^4$ increase drastically to a relatively higher value when Agents 1 and 4 are under attacks.
Therefore, it is beneficial to use the state estimation results to inform the controller about the presence of malicious attacks.
Although the performance index is insensitive to attacks, it never closely approaches one even when there is no attacks due to the inferior state estimation accuracy than external positioning systems.

\begin{figure}
   \centering
\hspace{-.1in}   \includegraphics[width=3.2in]{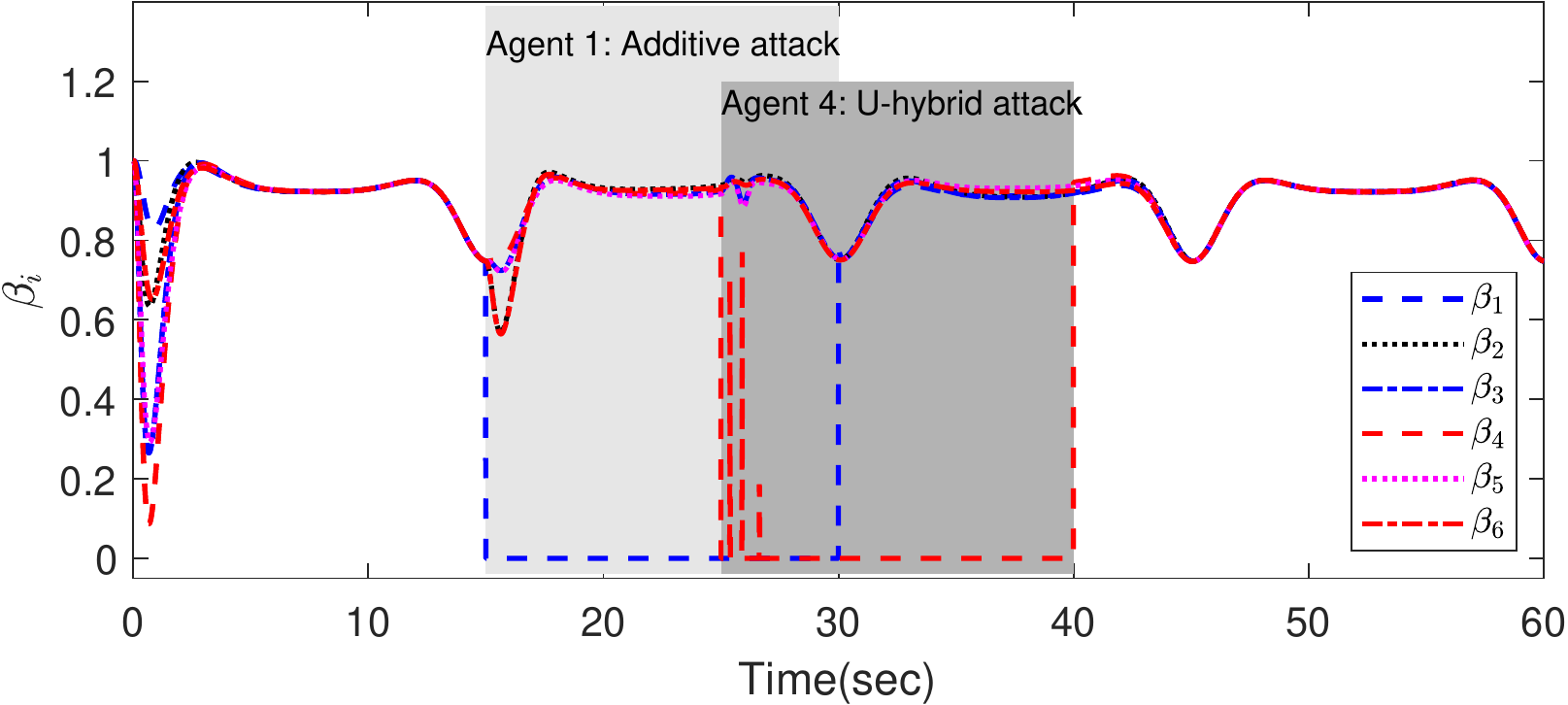}
   \caption{Evolution of the GPS quality measure $\beta_i$ of the mobile robot network under malicious attacks.}
\label{fig:sim:CLB:beta}
\end{figure}

\begin{figure}
   \centering
\hspace{-.1in}   \includegraphics[width=3.2in]{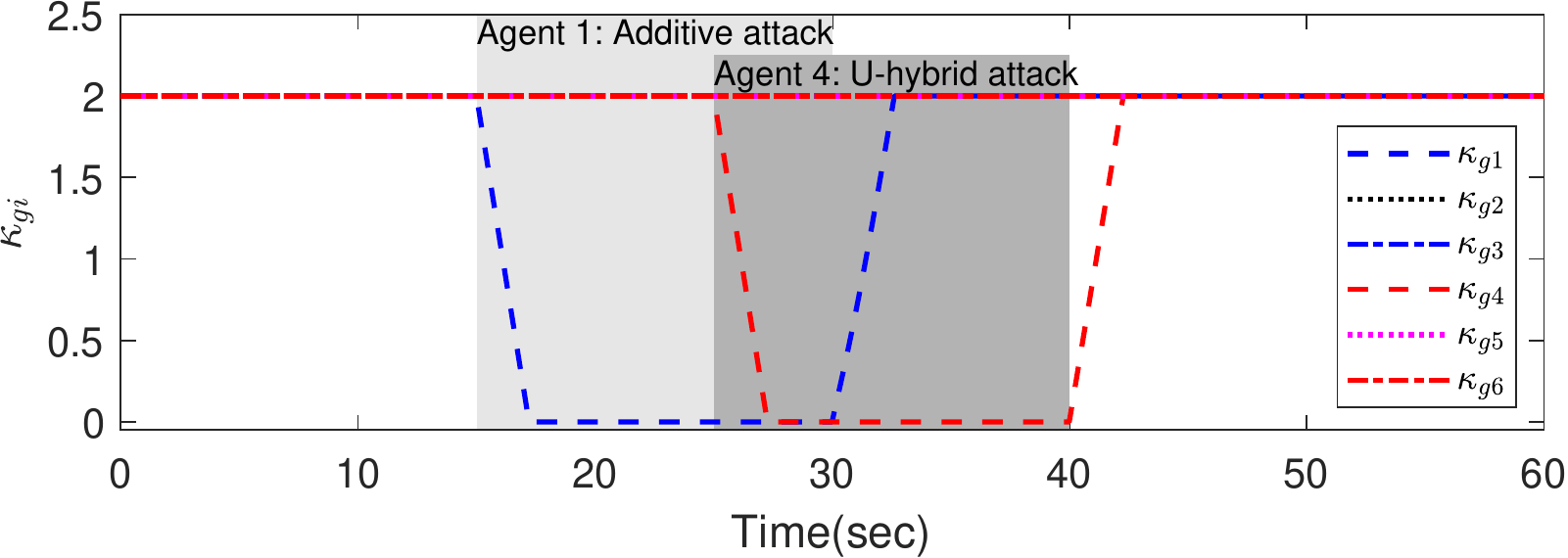}
   \caption{Evolution of the global tracking gains by using the proposed CL-based gain-tuning approach.}
\label{fig:sim:CLB:kappa}
\end{figure}

\subsubsection{CL-Based Gain-Tuning Approach}
To enhance the resilience and recovery performance of mobile robot network before, during, and after attacks, the CL recovery and gain-tuning approaches are combined to form the CL-based gain-tuning approach.
The GPS quality measure~\eqref{eq:beta}, obtained from KL divergence, is utilized to design a mechanism for regulating the global tracking gains $\kappa_{gi}$ in an adaptive fashion.
For $\sigma_{\beta_i}=3$ and $\chi_{\beta_i}=0.5$ with $\gamma_i=1$, the simulation results are illustrated in Figs.~\ref{fig:sim:resi:index},~\ref{fig:sim:CLB:beta}, and~\ref{fig:sim:CLB:kappa}.
It is noted that the CL-based gain-tuning approach is applied for all agents starting at $t=10$ sec after the system achieves asymptotic stability.
Fig.~\ref{fig:sim:CLB:beta} shows that the GPS quality measure decreases when the agent becomes under attack.
Therefore, the GPS quality measure can be used to indicate the health condition of the global positioning information.
For $\beta_i$ less than $\chi_{\beta_i}$, $\kappa_{gi}$ starts decreasing to zero so that the global positioning information subjected to malicious attacks can be excluded from the control system.
At $t=25$ sec, $\kappa_{g4}$ also decreases to zero because of the unstable attack on Agent 4 while other normal agents keep their own value.
%By comparing to Fig.~\ref{fig:sim:resi:gain}, only the agents under attack would have the global tracking gains decreasing to zero, and there is no influence to their adjacent agents.
Moreover, after the attacks are removed at $t=30$ sec and $t=40$ sec, respectively for Agent 1 and Agent 4, $\kappa_{g1}$ and $\kappa_{g24}$ can be recovered to the original value before attacks.

\subsubsection{Restoration}
\begin{comment}
\begin{figure}
   \centering
\hspace{-.1in}   \includegraphics[width=3.4in]{Figure/M_I_table}
   \caption{Performance indices for mobile robot network in various scenarios of multi-agent attacks by using CL-based gain-tuning approach. The attacks occur at $t_a=15$ sec without being removed.}
\label{fig:sim:kappag:comp}
\end{figure}
\end{comment}

%Among the aforementioned approaches, although CL-based gain-tuning can ensure a relatively higher performance index before, during, and after attacks, the performance drop during attacks is not negligible.
%Different sensory topology, control gain, type of attack, number of attack, and strength of attack would affect the resilience of the proposed approach.
During the simulation, the performance drop is an important indication for resilience.
We introduce the concept of restoration, $R_s$, as the reference to evaluate the recovery performance.
The restoration is defined by
\begin{eqnarray}\label{eq:def:RS}
&&\hspace{-.2in} R_s=\int_{t_a}^{t_r}\big(1-\mathcal I_f^g(\tau)\big)d\tau,
\end{eqnarray}
where $\mathcal I_f^g$ is the performance index defined in~\eqref{eq:perf:index}, $t_{a}$ is the time when an attack occurs, and $t_r$ is the instant when the performance index recovers to asymptotic stability that is $\mathcal I_f^g=1$.
The restoration can be considered as the size of the resilience triangle for the proposed networked robot systems for performance evaluations and comparisons.
In addition to $R_s$, the time-to-recover, $t_r-t_a$, and minimum $\mathcal I_g^f$ are also considered to evaluate the resilience of the system.

\section{Experimental Results}\label{sec:exp}
\begin{figure}
   \centering
\hspace{-.1in}   \includegraphics[width=3.35in]{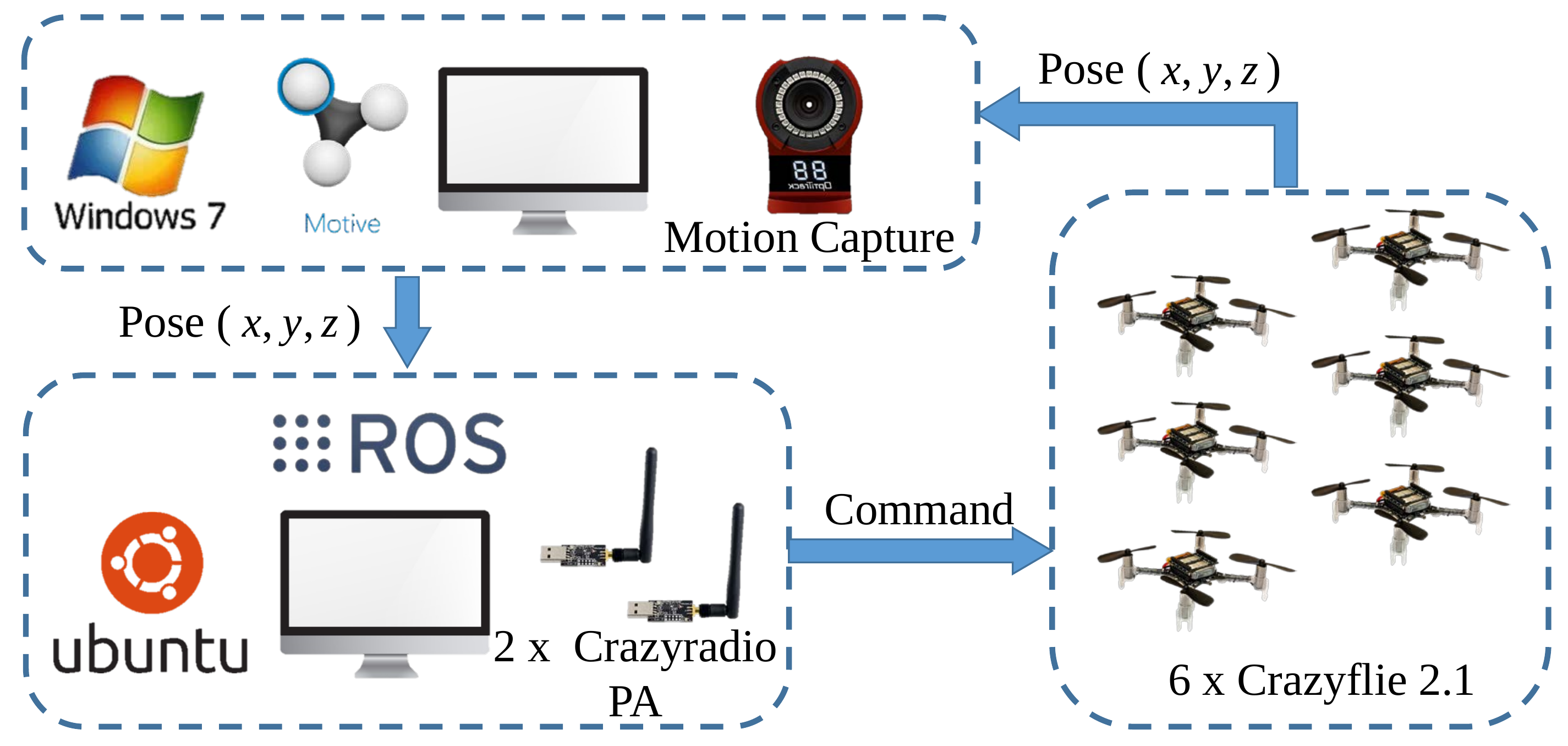}
   \caption{Experimental setup of the multi-quadrotor system for the implementation of the resilient TVFC in the presence of positioning attacks.}
\label{fig:exp:setup}
\end{figure}

In this section, we discuss the implementation of the proposed scheme on a group of six quadrotors that are subject to global positioning attacks.
%The implementation of the proposed resilience operation on time-varying formation tracking for mobile robot networks under positioning attacks is addressed in this section by using a group of six quadrotors.
It is noted that our general approach is not limited to flying robots.
%Nevertheless, the system of quadrotor flying robots is with highly potential in various applications and an emerging and significant topic in networked robot systems, so the proposed resilient TVFC systems are implemented in this paper on multiple quadrotor systems.
The experimental setup of the quadrotor network system is illustrated in Fig.~\ref{fig:exp:setup}.
Six Crazyflie 2.1, designed by Bitcraze, are used as the mobile agents in the mobile robot network, and the system is built upon the Robot Operating System (ROS) and its package  ``crazyswarm''~\cite{crazyswarm}. The motion of agents is captured by a motion capture (MOCAP) system consisted of IR cameras at 120 Hz, and then the commands from the proposed design are sent to agents at 100 Hz asynchronously.
\begin{comment}
Crazyflies have been used in many previous studies of multi-robot systems~\cite{araki2017multi, 8424034, 7797505}; their tiny size makes them suitable for indoor implementation.
In our experiments, a motion capture (MOCAP) system consisting of IR cameras (Optitrack Flex 13) is used to track the 3D positions of the flying agents within the testing area.
The motion tracking frequency is set at 120 Hz.
\end{comment}

\begin{comment}
In order to provide better computational efficiency, the MOCAP data-processing software, \textit{Motive}, runs on a separate PC.
The proposed control and estimation algorithms are implemented in the Robot Operating System (ROS) running on another Linux computer.
The proposed control and estimation system is integrated with the aforementioned hardware through ROS package  ``crazyswarm''~\cite{crazyswarm}, which is executed at 100 Hz.
The system collects the agents' position data from the MOCAP system and transmits the commands generated from the proposed controller to Crazyflies through radio modules called Crazyradio PA.
Note that both the global and relative positioning information are from the data of the MOCAP system containing noises from the hardware.
For emulating the global positioning attacks, corruption signals are artificially generated based on the corresponding localization information from the MOCAP system in ROS.
\end{comment}

% while implementing the control algorithms using global positioning information.

%Two scenarios are presented to conclude the effect of the proposed resilient control algorithm.
%For scenario 1, we focus on local formation tracking and the influence of different kinds of attacks.
%Further results with additional global trajectory tracking are demonstrated in scenario 2.

\subsection{Scenario 1: Time-Varying Formation with Stationary Global Trajectory}

\begin{figure}
   \centering
\hspace{-.1in}   \includegraphics[width=3.2in]{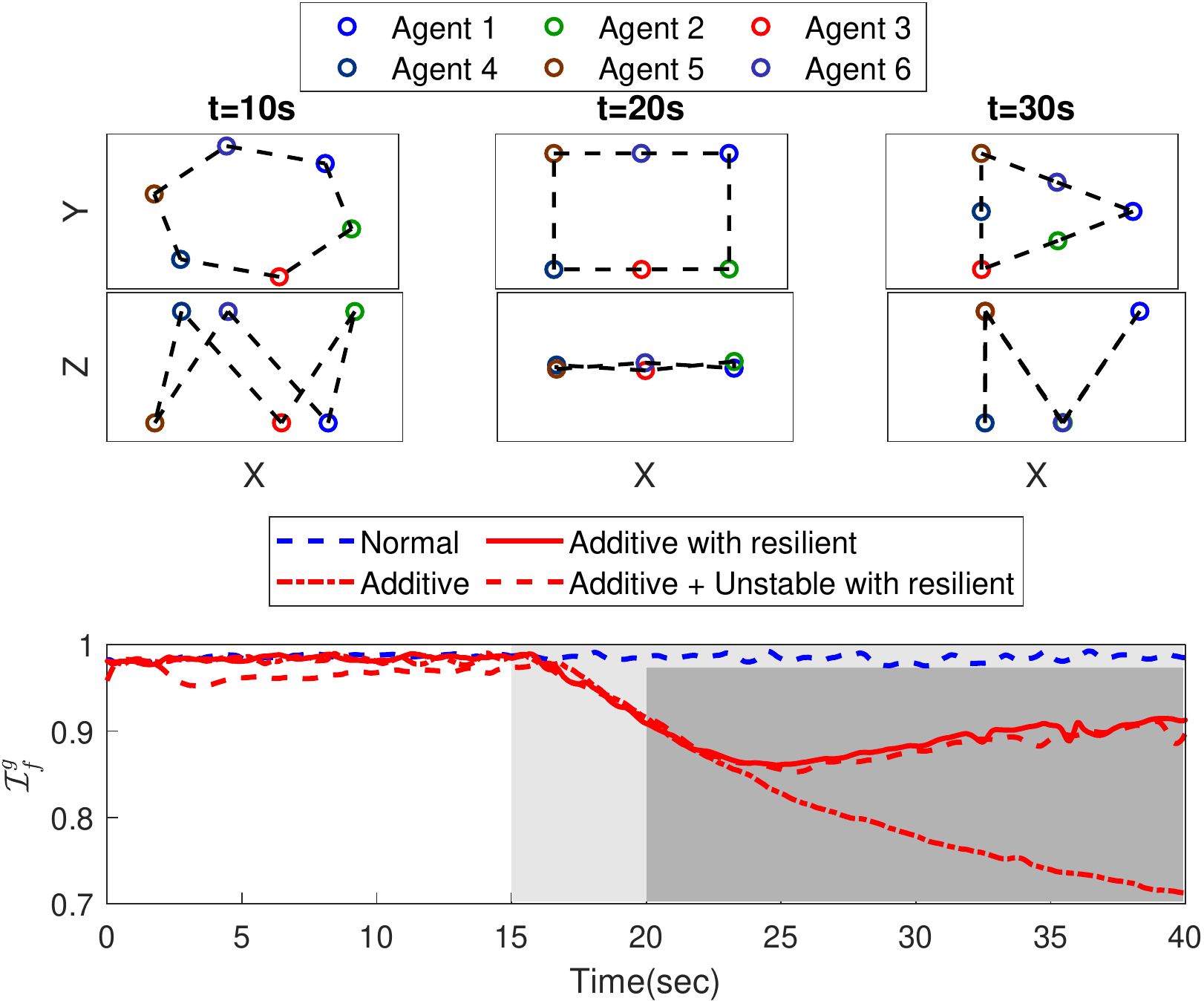}
   \caption{Desired formation (top) and performance indices (bottom) for the time-varying formation tracking with stationary global trajectory with/without global positioning attacks. For the cases with additive attack, Agent 1 is subject to attack for $t\geq t_{a1}=15$ sec. For the case with additional unstable attack, Agent 4 is under attack for $t\geq t_{a4}=20$ sec.}
\label{fig:exp:pi:noGL}
\end{figure}

To demonstrate the capability of the proposed resilient operations under positioning attacks, the time-varying formation tracking with stationary (time-invariant) global trajectory, $h^d = [0, 0, 0.9]^T$ m, is considered as the first scenario.
%By considering a stationary global trajectory, the influence from positioning attacks and recovery can be accessed separately without being affected by the time-dependency of the trajectory.
By considering a stationary global trajectory, the influence from positioning attacks and recovery can be accessed separately with time-invariant trajectories.
Six mobile agents arrive at horizontal formations of hexagon, rectangle, and triangle at $t=10$, $20$, and $30$ sec, respectively, as shown in Fig.~\ref{fig:exp:pi:noGL}.
The sensory topology of local positioning measures is represented by the dashed lines in each of the formations in Fig.~\ref{fig:exp:pi:noGL}.
Moreover, the vertical formations of the mobile agents are formed by trajectories of $\bar h_i^d$ that are given as $\bar h_{iz}^d(t)=0.15\sin(0.05\pi t+i\pi)$ m for $i=\{1,\ldots, 6\}$.

\begin{table}
\renewcommand\arraystretch{1.5}
\begin{center}
\small
\caption{Modified restoration ($\bar R_s^\text{exp}$) in experiments for the cases in Scenario 1 ($\int_{15}^{40}Nor(\mathcal I_f^g(\tau))d\tau=24.61$).}
\begin{tabular}{|c|c|c|}
 \hline
  {\bf Cases in Scenario 1} & \tabincell{c}{\bf Resilient Operation} & {\bf $\bar R_s^\text{exp}$} \\   \hline 
Normal (attack-free)  &  - & $0\%$ \\   \hline
Additive attack at $t_{a1}=15$ s   & No & $23.78\% $ \\   \hline
Additive attack at $t_{a1}=15$ s    &  Yes  & $8.48\%$ \\   \hline
\tabincell{c}{Additive attack at $t_{a1}=15$ s \\
Unstable attack at $t_{a4}=20$ s }  & Yes  & $8.99\%$ \\  \hline
%Agents $1\sim6$  & 20.42 & 0.42 & 2.65 \\  \hline
%Scenario 7  & 32.6 & 0.15 & 8.12 \\  \hline
%Scenario 8   & $\infty$ & 0  & $\infty$ \\  \hline
\end{tabular}\label{table:mod:restoration:exp1}
\end{center}
\end{table}

The tracking gains for the proposed TVFC are chosen as $\kappa_{gi}=0.8$, $\kappa_{fi}=0.4$, $\sigma_{fi}=0.05I_3$, and $w_{ij}=1$.
The performance indices of the robot network under different attack settings are illustrated in Fig.~\ref{fig:exp:pi:noGL}, where $\vartheta = 20$ and $\alpha_f^g = 3$ in $\mathcal I_f^g$.
%Four different cases are considered that are healthy positioning (normal), additive attack, additive attack with resilient operation, and additive plus unstable attacks with resilient operation.
With various positioning attacks, the quadrotors would deviate from their desired positions and cause degradation of the performance indices.
In the absence of malicious attack, the performance index $\mathcal I_f^g$ stably stays around $0.985$ providing satisfactory formation tracking.
When an additive attack is applied on Agent 1 with $\Delta_{xi}=[-1.5, -2.5, -1.0]^T$ m at $t_{a1}=15$ sec, the performance index, $\mathcal I_f^g$, degrades significantly without resilient operation as the tracking error of the compromised agent propagates through the network and causes its neighbors to drift away from their desired trajectories.
We then consider the same attack situation but now apply the proposed CL-based gain-tuning resilient operation~\eqref{eq:CL:gain:regu} with $\gamma_i = 1$, $\sigma_{\beta_i}=3$, and $\chi_{\beta_i}=0.5$.
It can be seen from Fig.~\ref{fig:exp:pi:noGL} that $\mathcal I_f^g$ would gradually increase and the performance improves significantly comparing to last case.
Here we chose $Q=0.02^2I_n$, $R^\text{GPS}=(5 \times 10^{-4})^2I_n$, $R^{r_{ij}}=(5 \times 10^{-4})^2I_n$, and $\chi = 4000$ based on the typical accuracy of the MOCAP system.
Furthermore, under an additional unstable attack on Agent 4 with $\delta_{x4}=1$ and $c_{a4}=5$ at $t_{a4}=20$ sec, the robot team maintains a better performance, and $\mathcal I_f^g$ is larger than $0.85$ and keeps increasing for better formation tracking.

To quantitatively assess the performance of the system in experiments, we define a new measure modified from the restoration $R_s$ in~\eqref{eq:def:RS} as
\begin{eqnarray}\label{eq:def:RS:exp}
&&\hspace{-.2in} \bar R_s^\text{exp}=\frac{\int_{t_a}^{t_s}\big(Nor(\mathcal I_f^g)-Att(\mathcal I_f^g)\big)d\tau}{\int_{t_a}^{t_s}Nor(\mathcal I_f^g)d\tau},
\end{eqnarray}
where $t_s$ is the terminal time of the experiment, $Nor(\mathcal I_f^g(t))$ is the reference of the performance index without attacks, and $Att(\mathcal I_f^g(t))$ is the performance index under attacks with/without resilience.
Since $Nor(\mathcal I_f^g(t))$ is not equal to one, the modified restoration, $\bar R_s^\text{exp}$, provides a way to compare the resilience performance of the system in experiments to the cases without attacks. %is defined for experiment to express the resilience performance comparing to the case of the networked robotic system without attack.
It can be seen as the degree of performance degradation from the attack-free case, with $0\%$ representing no degradation and $100\%$ being full degradation.
Table~\ref{table:mod:restoration:exp1} summarizes the modified restoration of different cases in Scenario 1.
%The modified restoration with the proposed resilient operation is much lower than the case where the mobile robot network is under additive attack but without resilient operation.
%Even though there are multiple attacks on the mobile agents, the resilient operation is able to maintain a relatively high performance index in the time-varying formation tracking control.
Obviously, the proposed resilient operation holds better  better performance even if there is unstable attack.

\subsection{Scenario 2: Time-Varying Formation and Global Trajectory}

\begin{figure}
   \centering
\hspace{-.1in}   \includegraphics[width=3.2in]{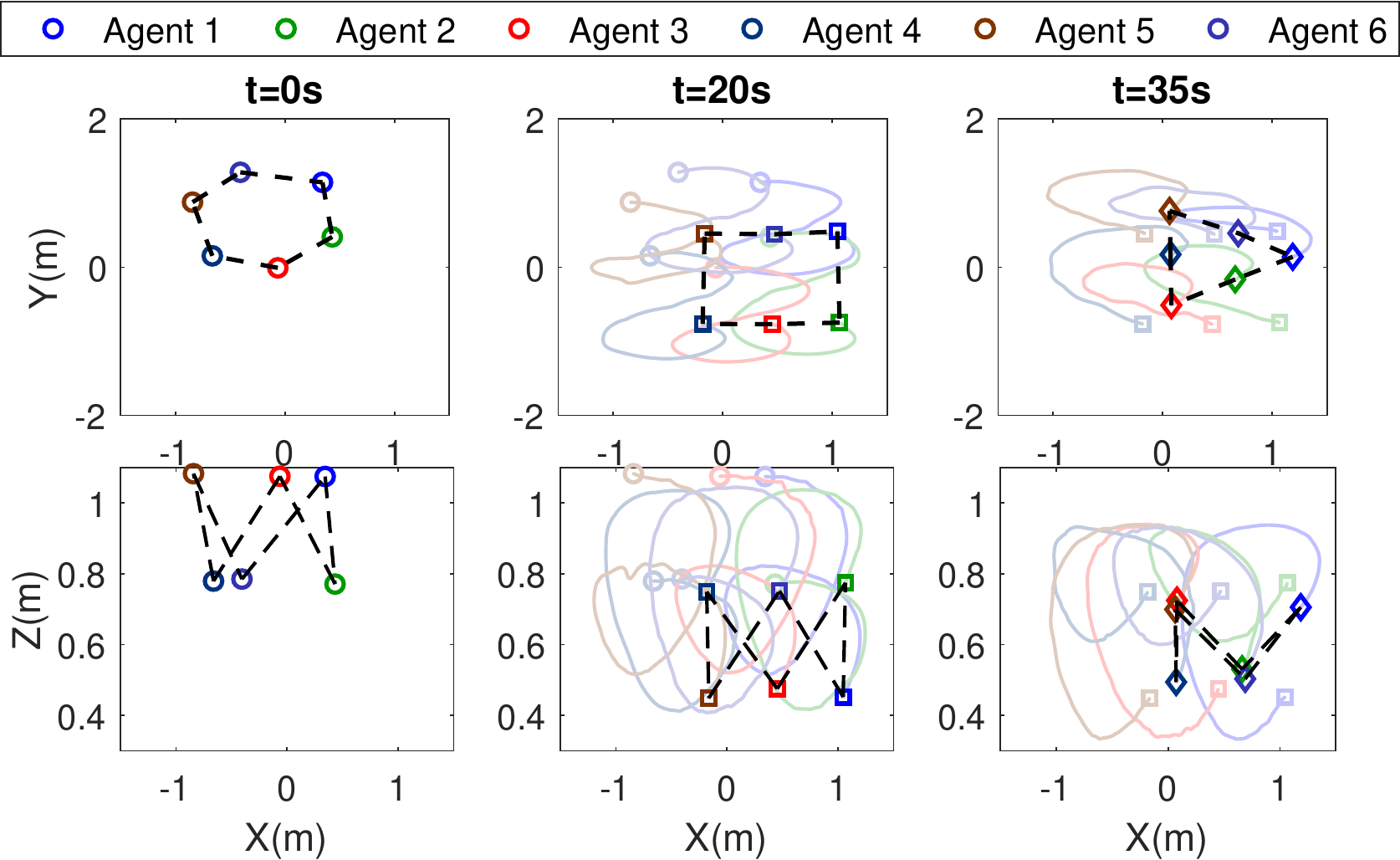}
   \caption{Snapshots of the time-varying formation and global trajectory tracking in the absence of global positioning attacks.}
\label{fig:exp:traj:GL}
\end{figure}

To further investigate the effect of the proposed TVFC for mobile robot networks and the resilient operation, two advanced cases are presented where the mobile robot network tracks a predefined global trajectory. %with tracking a predefined global trajectory are presented.
The quadrotor team reaches hexagon, rectangle, and triangle horizontal formations at $t=5$, 20, and 35 sec, respectively.
The vertical formation is the same as in the previous case.
Additionally, the global trajectory of the mobile agents is defined as $h^d=[1.3\sin(2\pi t/15)/(\cos(2\pi t/15)-3),-1.2\cos(\pi t/15)/(\cos(2\pi t/15)-3),0.2\cos(2\pi t/15)+0.7]^T$ m, which is a 3D lemniscate trajectory with a time-varying $Z$-component.
The control gains are chosen as $\kappa_{gi}=0.4$, $\kappa_{fi}=0.4$, and $\sigma_{fi}=0.1I_3$.

\begin{comment}
\begin{table}
\renewcommand\arraystretch{1.5}
\begin{center}
\small
\caption{Modified restoration ($\bar R_s^\text{exp}$) of the mobile robot network in experiments under different attacks with/without the proposed resilient operation in Scenario 2 ($\int_{15}^{40}Nor(\mathcal I_f^g(\tau))d\tau=23.03$).}
\begin{tabular}{|c|c|c|}
 \hline
  {\bf Cases in Scenario 2} & \tabincell{c}{\bf Resilient \\ \bf Operation} & {\bf $\bar R_s^\text{exp}$} \\   \hline 
Normal (attack-free)  &  - & $0\%$ \\   \hline
\tabincell{c}{Additive Attack $t_{a1}=15$ sec \\
U-Hybrid Attack $t_{a4}=20$ sec }  & Yes  & $1.98\%$ \\  \hline
%Agents $1\sim6$  & 20.42 & 0.42 & 2.65 \\  \hline
%Scenario 7  & 32.6 & 0.15 & 8.12 \\  \hline
%Scenario 8   & $\infty$ & 0  & $\infty$ \\  \hline
\end{tabular}\label{table:mod:restoration:exp2}
\end{center}
\end{table}
\end{comment}

\begin{figure}
   \centering
\hspace{-.1in}   \includegraphics[width=3.2in]{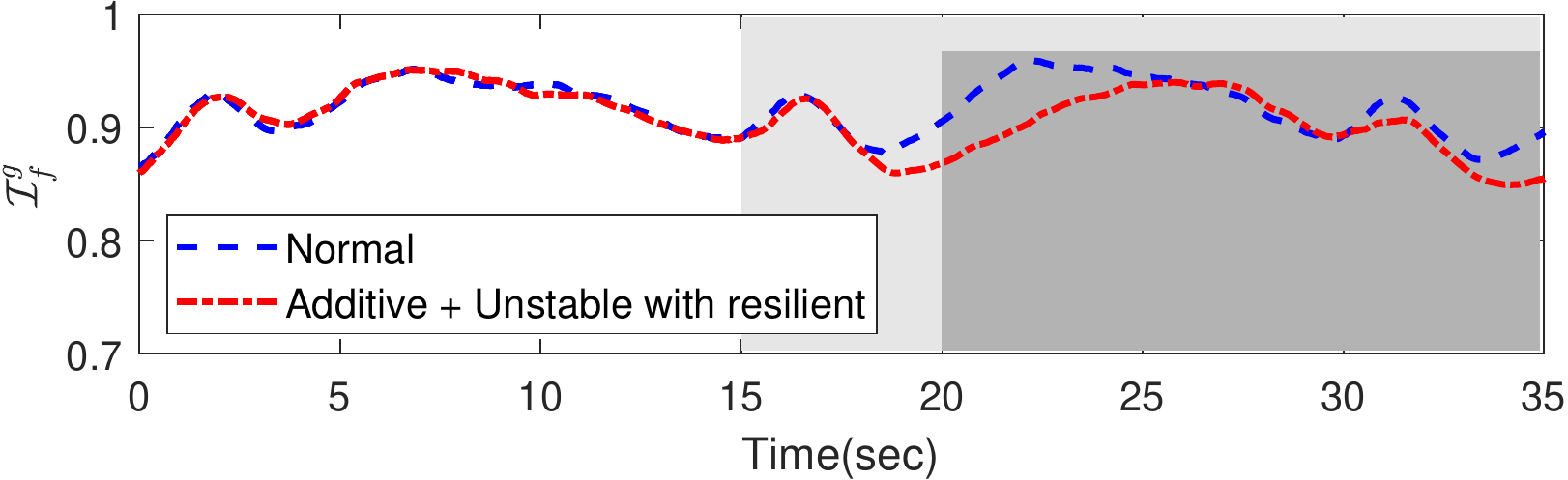}
   \caption{Performance indices of the experiments with healthy global positioning, and under additive and unstable attacks with resilient operation. For the case with attacks, Agent 1 is subject to additive attack for $t\geq t_{a1}= 15$ sec, and Agent 4 is under unstable attack for $t\geq t_{a4}=20$ sec.}
\label{fig:exp:pi:GL}
\end{figure}

The snapshots of the formation tracking without attacks are shown in Fig.~\ref{fig:exp:traj:GL}, where the dashed lines represent the local measurement network among the mobile agents.
The performance indices of the proposed control algorithm tracking the global and local trajectories with and without global positioning attacks are shown in Fig.~\ref{fig:exp:pi:GL}.
In the absence of attacks, the performance index varies around $0.9$, indicating that the mobile robot network can closely track the global trajectory and time-varying formation.
%The modified restoration of the experimental results in Scenario 2 is illustrated in Table~\ref{table:mod:restoration:exp2}.
%It is shown that, with the global time-varying trajectory tracking, the resilient operation can ensure satisfactory functionality by providing superior performance close to the case without malicious attacks.
%In this case, $\bar R_s^\text{exp}$ is better than Scenario 1 because $Nor(\mathcal I_f^g)$ is relative smaller in the case with the  global time-varying trajectory.

\begin{figure}
   \centering
\hspace{-.1in}   \includegraphics[width=3.2in]{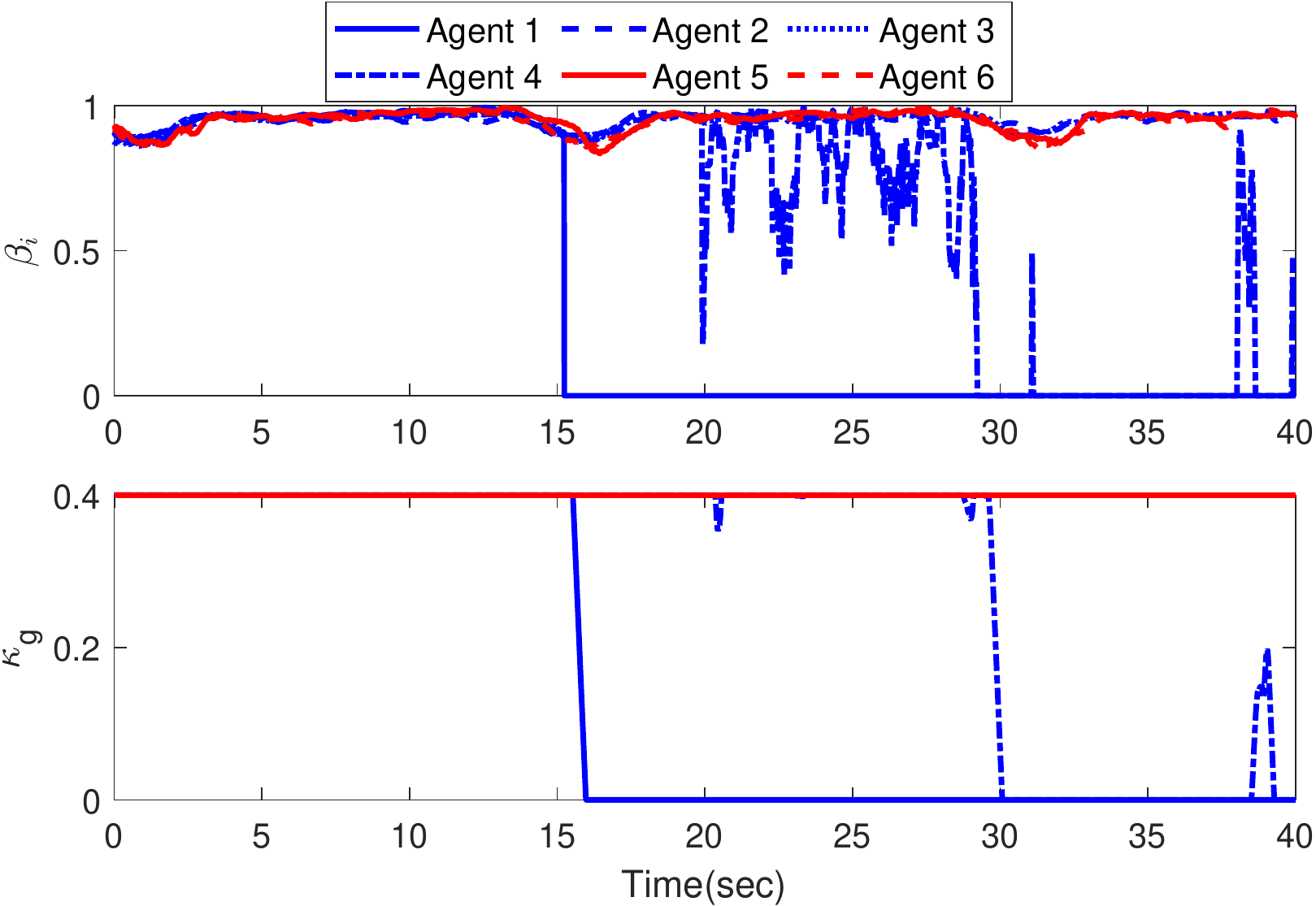}
   \caption{Evolution of $\beta$ and $\kappa_g$ in the experimental case with both additive attack on Agent 1 and unstable attack on Agent 4. Parameter projection~\cite{Dixon07TAC,Liu15TMECH} is utilized in~\eqref{eq:CL:gain:regu} such that $0\leq \kappa_{gi}\leq 0.4$.}
\label{fig:exp:gain:GL}
\end{figure}

When Agent 1 and Agent 4 are under additive and unstable attacks, respectively, the performance index slightly degrades due to the attacks but the tracking performance remains as satisfactory as the case without attack in Fig.~\ref{fig:exp:pi:GL}.
The parameters used in cooperative localization are identical to the previous section with the stationary global trajectory.
This result implies that the proposed control algorithm is capable of eliminating severe influence of malicious attacks on agents to protect the systems from attacks and recovery to attack-free performance.
The evolution of $\beta_i$ and $\kappa_{gi}$ is shown in Fig.~\ref{fig:exp:gain:GL}, where the identification of attacks is via $\beta_i$ and the corresponding global tracking gains decrease to eliminate the adversarial effect on the mobile robot network.

%%%%%%%%%%%%%%%%%%%%%%%%%%%%%%%%%%%%%%%%%%%%%%%%%%%%%%%%%%%%%%%%%%
\section{Conclusion}\label{sec:conclu}

The persistency of maintaining a time-varying formation to track the reference trajectory of a mobile robot network under global positioning attacks is investigated in this paper. Based on the local and global positioning information, control algorithms are presented and studied to ensure formation tracking performance. Subsequently, Lyapunov-based stability analyses are provided in the presence of three types of malicious attacks on the global positioning signals that are additive attack, hybrid attack, and unstable attack. Moreover, a performance index is presented to evaluate the efficiency of mobile robot networks under time-varying formation tracking control with/without compromised agents subjected to adverse impacts. Simulation results and experiments validate the control system and the influence of adversarial attacks on global positioning.
Future work will address the resilience in coordinating heterogeneous robotic systems, reaction to sensory/actuator faults, and handling unreliable sensory topologies.

\appendices
\renewcommand{\theequation}{\thesection.\arabic{equation}}
\section{Proof for Theorem~\ref{thm:global}}
\setcounter{equation}{0} % reset equation number counter 
\begin{proof}\label{proof:thm:global}

Without loss of generality, let us consider the Lyapunov function candidate for the $i^{th}$ robot in the the mobile robot network as $V_i(\xi_i)=\frac{1}{2}\xi_i^T\sigma_{fi}^{-1}\xi_i$.
By taking the time-derivative of $V_i$ along the trajectories of~\eqref{eq:cls:xi}, we obtain $\dot V_i=-\xi_i^T\xi_i$, which is negative definite.
Since $V_i$ is positive definite and $\dot V_i$ is negative definite, we obtain $V_i(\xi_i(t))\leq V_i(\xi_i(0))$ so that $\xi_i\in\mathcal L_\infty$ because $\sigma_{fi}$ is a positive-definite matrix.
Next, by integrating $\dot V_i(t)$ with respect to time from $0$ to $\infty$, we have $V_i(t)-V_i(0)=-\int_0^t\xi_i^T(\tau)\xi_i(\tau)d(\tau)$, which results in $\int_0^t\xi_i^T(\tau)\xi_i(\tau)d(\tau)+V_i(t)=V_i(0)$.
Since $V_i(\xi_i)$ is positive definite, we can further obtain that $\int_0^t\xi_i^T(\tau)\xi_i(\tau)d(\tau)\leq V_i(0)<\infty$; consequently, we have $\xi_i\in\mathcal L_2$.
From the closed-loop dynamics~\eqref{eq:cls:xi}, we further obtain that $\dot\xi_i\in\mathcal L_\infty$.
Since $\xi_i\in\mathcal L_2 \cap \mathcal L_\infty$ and $\dot\xi_i\in\mathcal L_\infty$, we get from Barbalat's lemma~\cite{khalil} that $\lim_{t\rightarrow\infty}\xi_i(t)=0$.
Therefore, we conclude that $\xi_i$ is asymptotically stable.
%Consequently, we conclude that $\lim_{t\rightarrow\infty}\xi_i(t)=0$ from $\xi_i\in\mathcal L_2 \cap \mathcal L_\infty$.

Next, let us rewrite $\xi_i=\dot{\tilde x}_i+\kappa_{gi}\tilde x_i+\kappa_fe_i$ as $\dot{\tilde x}_i=-\kappa_{gi}\tilde x_i-\kappa_f e_i+\xi_i$, which can be considered as the state equation of $\tilde x_i$ with $\xi_i$ as the input.
From the definition of $e_i$ with the weighted Laplacian matrix, we have $e=[L\otimes I_n]\tilde x$, where $e=[e_1^T,\ldots, e_N^T]^T\in \mathbb R^{Nn}$ and $\tilde x=[\tilde x_1^T, \ldots, \tilde x_N^T]^T\in \mathbb R^{Nn}$.
By considering $\xi\in\mathbb R^{Nn}$ as the stacked vector of $\xi_i$ such that $\xi=[\xi_1^T, \xi_2^T, \ldots, \xi_N^T]^T$, the interconnected system becomes
\begin{align}\label{eq:dxi:cls}
\dot{\tilde x}=&-D_{\kappa_{g}}\otimes I_n\tilde x-\kappa_f[L\otimes I_n]\tilde x+\xi\nonumber \\
=&-[(D_{\kappa_{g}}+\kappa_fL)\otimes I_n]\tilde x+\xi,
\end{align}
where $D_{\kappa_{g}}\in\mathbb R^{N\times N}$ denotes a diagonal matrix with the $i^{th}$ diagonal term as $\kappa_{g}$.
Since $\kappa_f$, $\kappa_{gi}$ are all positive, we can show that $-(D_{\kappa_{g}}+\kappa_fL)$ is a Hurwitz matrix~\cite{khalil} because $L(\mathcal G)$ has an isolated eigenvalue of zero and all others with a positive real part (Property~\ref{pro:L:eig}).
As $\xi$ converges to the origin as time goes to infinity, the linear time-invariant (LTI) system~\eqref{eq:dxi:cls} with $\xi$ as the input is asymptotically stable.
Consequently, $\tilde x, \dot{\tilde x}\in\mathcal L_2 \cap \mathcal L_\infty$ so that $\tilde x\rightarrow 0$ as $t\rightarrow \infty$.
Furthermore, by observing~\eqref{eq:dxi:cls}, the convergence of $\tilde x$ and $\xi$ to the origin lead to $\dot{\tilde x}\rightarrow 0$ asymptotically.
Consequently, $\lim_{t\rightarrow\infty}\tilde x_i(t)=0$ and $\lim_{t\rightarrow\infty}\dot{\tilde x}_i(t)=0$ demonstrate that $\lim_{t\rightarrow\infty}\left(x_i(t)-h_i^d(t)\right)=0$ and $\lim_{t\rightarrow\infty}\dot x_i(t)=\lim_{t\rightarrow\infty}\dot h_i^d(t)$.
The mobile robot network achieves global formation tracking.\qedhere

\end{proof}

\section{Proof for Theorem \ref{thm:estimator_consistency}}
\setcounter{equation}{0} % reset equation number counter 
\begin{proof}\label{proof:thm:estimator_consistency}

    Based on measurement model \eqref{eq:update_r}, the measurement error, or innovation, of the $i^{th}$ agent relative to the $j^{th}$ agent can be calculated as 
    \begin{align}
        \tilde{s}^{r_{ij}}_i(k) =& s^{r_{ij}}_i(k) - \hat{s}^{r_{ij}}_i(k) \nonumber \\
        =& x_j(k) - x_i(k) + \omega^{r_{ij}}_i(k) - \left[ h^d_j(k) - \hat{x}_i(k|k-1) \right] \nonumber \\
        = & \tilde{x}_j(k) - \tilde{x}_i(k|k-1) + \omega^{r_{ij}}_i(k).
    \end{align}
    Recall that $\tilde{x}_j(k)$ denotes the global tracking error. Based on the definition for information vector, $H^{r_{ij}}_i = -I_{3}$, and \eqref{eq:infovec_update}, the  state estimation error after update with inter-agent measurement can be calculated as 
    \begin{align}
        \tilde{x}_i(k|k) =& x_i(k) - \hat{x}_i(k|k) \nonumber \\
        =&x_i(k) -\Phi_i(k|k)^{-1}\varphi_i(k|k) \nonumber \\
        =&x_i(k) - \Phi_i(k|k)^{-1}\Big\{\varphi_i(k|k-1) \nonumber \\
        & \qquad -R_i(k)^{-1}\big[\tilde{s}_i^r(k) - \hat{x}_i(k|k-1) \big] \Big\}. \label{eq:est_error2}
    \end{align}
    In this proof, we use a simplified notation for the inter-agent measurement covariance matrix: $R_i(k) = R^{r_{ij}}_i(k)$. Left multiplying $\Phi_i(k|k)$ on both sides of \eqref{eq:est_error2} and rearranging yield
    \begin{align}
        &\Phi_i(k|k)\tilde{x}_i(k|k) \nonumber \\
        &= \Phi_i(k|k-1)\left[x_i(k)-\hat{x}_i(k|k-1)\right] \nonumber \\
        & \qquad \qquad \qquad \quad + R_i(k)^{-1}\left[\tilde{s}_i^{r_{ij}}(k) + x_i(k) - \hat{x}_i(k|k-1)\right] \nonumber \\
        &=\Phi_i(k|k-1)\tilde{x}_i(k|k-1) + R_i(k)^{-1}\left[\tilde{s}_i^{r_{ij}}(k)+\tilde{x}_i(k|k-1)\right] \nonumber \\
        &= \Phi_i(k|k-1)\tilde{x}_i(k|k-1) + R_i(k)^{-1}\left[\tilde{x}_j(k) + \omega^{r_{ij}}_i(k)\right],
    \end{align}
    which leads to $\tilde{x}_i(k|k) = \Phi_i(k|k)^{-1}\Big\{\Phi_i(k|k-1)\tilde{x}_i(k|k-1) $
     $+ R_i(k)^{-1}\left[\tilde{x}_j(k)  + \omega^{r_{ij}}_i(k)\right]\Big\}$.
    Taking the expectation of both sides and the limit when $k\rightarrow \infty$ leads to
    \begin{align}
        E\left[\tilde{x}_i(k|k)\right] = \Phi_i(k|k)^{-1}\Phi_i(k|k-1)E\left[\tilde{x}_i(k|k-1)\right]. \label{eq:est_err_exp}
    \end{align}
    Here we applied $\lim_{k\rightarrow\infty} \tilde{x}_j(k) = 0$ and $E[\omega^{r_{ij}}_i(k)] = 0$.

    To show that $\tilde{x}_i(k|k)$ strictly decreases after the inter-agent measurement update, we show that the scalar-valued quadratic function 
    $
        V(k|k):= E[\tilde{x}_i(k|k)]^T\Phi_i(k|k)E[\tilde{x}_i(k|k)]
    $ 
    decreases from $V(k|k-1)$. Using the result from \eqref{eq:est_err_exp}, we can find that
    \begin{align}
        % V(k|k) = & E[\tilde{x}_i(k|k-1)]^T\Phi_i(k|k-1)^T(\Phi_i(k|k)^{-1})^T \Phi_i(k|k) \nonumber \\
        % & \qquad \qquad \Phi_i(k|k)^{-1}\Phi_i(k|k-1)E[\tilde{x}_i(k|k-1)] \nonumber \\
        % =& E[\tilde{x}_i(k|k-1)]^T\Phi_i(k|k-1)\Phi_i(k|k)^{-1} \nonumber \\
        % & \qquad \qquad \qquad \Phi_i(k|k-1)E[\tilde{x}_i(k|k-1)]. \label{eq:lyapunov}
        V(k|k) = & E[\tilde{x}_i(k|k-1)]^T\Phi_i(k|k-1)^T\Phi_i(k|k)^{-1} \nonumber \\
        & \qquad \qquad \qquad \Phi_i(k|k-1)E[\tilde{x}_i(k|k-1)]. \label{eq:lyapunov}
    \end{align}
    By applying the Woodbury matrix identity, we can find that 
    \begin{align}
         \Phi_i(k|k)^{-1} = & [\Phi_i(k|k-1) + R_i(k)^{-1}]^{-1} \nonumber \\
        = &\Phi_i(k|k-1)^{-1} - [\Phi_i(k|k-1) \nonumber \\
        & + \Phi_i(k|k-1)R_i(k)\Phi_i(k|k-1)]^{-1}. \label{eq:woodbury}
    \end{align}
    Substituting \eqref{eq:woodbury} into \eqref{eq:lyapunov} and rearranging lead to
    \begin{align}
        V(k|k) = & E[\tilde{x}_i(k|k-1)]^T\Phi_i(k|k-1)^TE[\tilde{x}_i(k|k-1)] \nonumber \\
        &- E[\tilde{x}_i(k|k-1)]^T\Phi_i(k|k-1)^T [\Phi_i(k|k-1) \nonumber \\
        & + \Phi_i(k|k-1)R_i(k)\Phi_i(k|k-1)]^{-1} \nonumber \\
        & \Phi_i(k|k-1)E[\tilde{x}_i(k|k-1)].
    \end{align}
    Note that the first term on the right-hand side is $V(k|k-1)$, and the second term is non-positive, i.e., $V(k|k) \leq V(k|k-1)$. This proves that $V(k|k)$ decreases from $V(k|k-1)$ after inter-agent measurement update.\qedhere

\end{proof}
% or


%\section{}


%\section*{Acknowledgment}
%The authors would like to thank Shih-Hsiu Chen and Mu-Tai Lin for providing the simulation and experiment data.

%The authors would like to thank the Technical Editor and the anonymous reviewers for their valuable and helpful suggestions that led to significant improvements of this paper.

\ifCLASSOPTIONcaptionsoff
  \newpage
\fi



%\bibliographystyle{IEEEtran}

\bibliographystyle{IEEEtran}

\bibliography{Harvey-V40, song, CF}

%\begin{IEEEbiography}{Yen-Chen Liu}
%\begin{IEEEbiography}[{\includegraphics[width=1in,height=1.25in,clip,keepaspectratio]{Bio_Liu}}]{Yen-Chen Liu}
%received the B.S. and M.S. degrees in Mechanical Engineering from the National Chiao Tung University, Hsinchu, Taiwan in 2003 and 2005, respectively, and the Ph.D. degree in Mechanical Engineering from University of Maryland, College Park in 2012.

%He is currently an Associate Professor in the Department of Mechanical Engineering, National Cheng Kung University, Tainan, Taiwan. His research interests include control of robotic system, bilateral teleoperation, multi-robot system, semi-autonomous system, and human-robot interaction.
%\end{IEEEbiography}

\end{document}